\documentclass[11pt,a4paper]{article}
\usepackage{times,latexsym}
\usepackage{url}
\usepackage[T1]{fontenc}

\usepackage[acceptedWithA]{tacl2021v1}

\usepackage{xspace,mfirstuc,tabulary}

\newif\iftaclinstructions
\taclinstructionsfalse %
\iftaclinstructions
\renewcommand{\confidential}{}
\renewcommand{\anonsubtext}{(No author info supplied here, for consistency with
TACL-submission anonymization requirements)}
\newcommand{\instr}
\fi

\iftaclpubformat %

\else

\fi

\usepackage{amsmath,amsfonts,bm,mathtools,amssymb}
\usepackage{booktabs}
\usepackage{graphicx}
\usepackage{multirow,longtable}
\usepackage{nccmath}
\usepackage{makecell}
\usepackage{footmisc}
\usepackage{cleveref}
\usepackage{xcolor,colortbl}
\usepackage{subcaption}  %
\usepackage{multirow, makecell}

\title{Large Language Models Are Human-Like Internally}

\author{Tatsuki Kuribayashi${}^{1,4}$\, Yohei Oseki${}^2$\, Souhaib Ben Taieb${}^{1, 3}$ \\ \textbf{Kentaro Inui${}^{1,4,5}$} \, \textbf{Timothy Baldwin${}^{1,6}$} \\
        ${}^{1}$MBZUAI
        ${}^{2}$The University of Tokyo  \,
        ${}^{3}$University of Mons  \\
        ${}^{4}$Tohoku University \,
        ${}^{5}$RIKEN  \,
        ${}^{6}$The University of Melbourne \\
  \texttt{\{tatsuki.kuribayashi,souhaib.bentaieb,}\\ \texttt{kentaro.inui,timothy.baldwin\}@mbzuai.ac.ae} \\
  \texttt{oseki@g.ecc.u-tokyo.ac.jp}
}

\begin{document} 
\maketitle
\begin{abstract}

Recent cognitive modeling studies have reported that larger language models (LMs) exhibit a poorer fit to human reading behavior~\cite{Oh2023-zw,Shain2022-qv,kuribayashi-etal-2024-psychometric}, leading to claims of their cognitive implausibility. In this paper, we revisit this argument through the lens of mechanistic interpretability and argue that prior conclusions were skewed by an exclusive focus on the final layers of LMs. Our analysis reveals that next-word probabilities derived from internal layers of larger LMs align with human sentence processing data as well as, or better than, those from smaller LMs. This alignment holds consistently across behavioral (self-paced reading times, gaze durations, MAZE task processing times) and neurophysiological (N400 brain potentials) measures, challenging earlier mixed results and suggesting that the cognitive plausibility of larger LMs has been underestimated. Furthermore, we first identify an intriguing relationship between LM layers and human measures: earlier layers correspond more closely with fast gaze durations, while later layers better align with relatively slower signals such as N400 potentials and MAZE processing times.
Our work opens new avenues for interdisciplinary research at the intersection of mechanistic interpretability and cognitive modeling.\footnote{Code is available at \url{https://github.com/kuribayashi4/surprisal_internal_layers}}

\end{abstract}

\section{Introduction}
\label{sec:intro}

Understanding human sentence processing has long been a fundamental goal in linguistics. This goal is typically approached by investigating \textit{what computational models can simulate human sentence processing data}, such as eye movement patterns during reading, in the field of computational psycholinguistics~\cite{Crocker2010-cp,Beinborn2024-lw}.  Natural language processing (NLP) models, such as neural language models (LMs), have played a crucial role in this endeavor, serving as tools to test linguistic hypotheses.  Specifically, the theory of expectation-based human sentence processing~\cite{hale-2001-probabilistic,Levy2008Expectation-basedComprehension,Smith2013-ap} --- which posits that humans continuously predict upcoming linguistic information during reading --- naturally raises the following questions:  how well do word probabilities (i.e., surprisal, $-\log p(\mathrm{word}|\mathrm{context})$) derived from LMs align with human sentence processing behavior?  What kind of LMs produce the most human-like surprisal?

Previous studies have provided substantial evidence supporting expectation-based accounts of human sentence processing (\citealt{Shain2022-qv}; \textit{inter alia}).  However, they reveal an intriguing trend: surprisal estimates from large language models (LLMs) often deviate from human reading behavior, and rather smaller models, such as GPT-2 small, offer better simulations of human behavior~\cite{Shain2022-qv,Oh2023-zw,kuribayashi-etal-2022-context,kuribayashi-etal-2024-psychometric}.
This observation --- \textit{larger LMs are less human-like} --- has sparked intriguing linguistic questions~\cite{Wilcox2024-qx} as well as a fair amount of confusion within the community. 
Why do smaller LMs appear more human-like, despite their generally poorer linguistic competence~\cite{Waldis2024-rf}? 

\begin{figure}[t]
    \centering
    \includegraphics[width=1.0\linewidth]{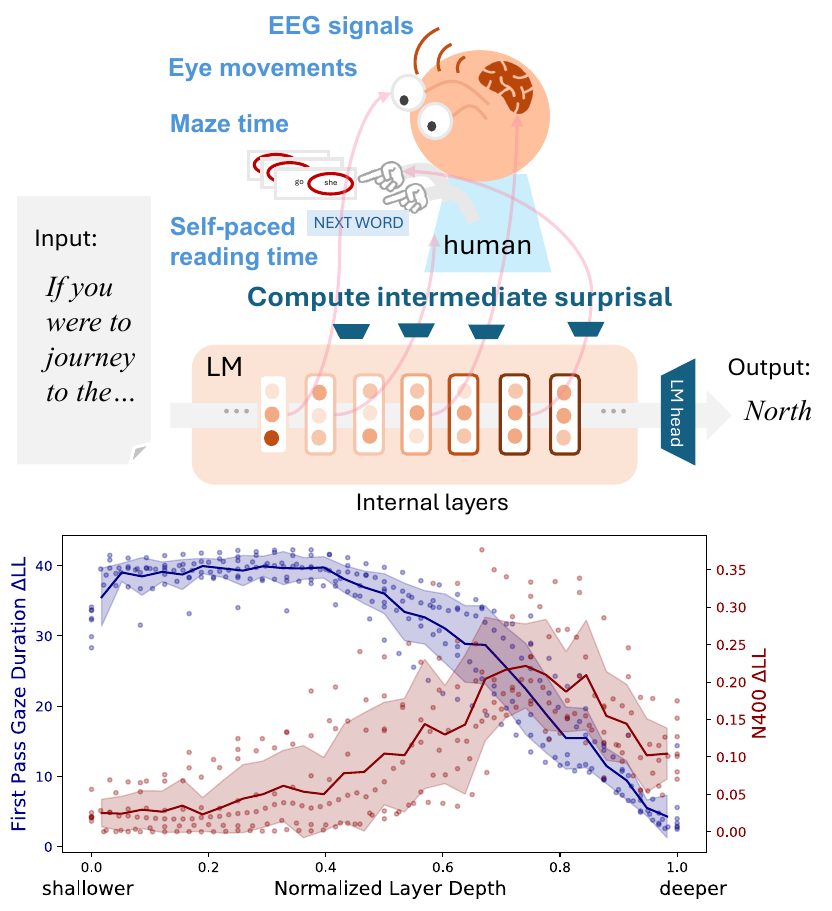}
    \caption{Different measures of human sentence processing align with surprisal from different layers of language models (LMs), and the best layer is typically not the final layer. 
    In the bottom plot, for example, gaze duration (blue dots) and EEG signal (red dots) correlate with earlier and later layers of LMs, respectively. Each dot corresponds to the fit of surprisal (y-axis) from a particular layer depth (x-axis) to human data (ZuCO corpus).
    }
    \label{fig:overview}
\end{figure}

In this work, we highlight the cognitively \textit{plausible} aspects of LLMs, challenging existing conclusions. 
Specifically, we show that \textbf{surprisal derived from the internal layers of larger LMs aligns with human sentence processing data as well as, or even better than, that from smaller LMs}. 
Previous studies, focusing exclusively on final-layers' surprisal, have overlooked this critical insight. 
These results could be drawn with techniques from mechanistic interpretability~\cite{dar-etal-2023-analyzing,belrose2023eliciting,Wendler2024-wr}, \textit{logit lens}~\cite{logitlens} or dubbed \textit{early exits}~\cite{icml19shallowdeepnetworks}; we compute next-word surprisals directly from internal layers of LMs by projecting intermediate representations into the output vocabulary space, bypassing subsequent layers. 
We additionally reveal that surprisal from earlier layers fits better with fast human responses (first-pass gaze durations and self-paced reading time), while surprisal from later layers aligns more closely with slower measures (N400 and MAZE task data) (Figure~\ref{fig:overview}). 
This also resolves the previously suggested \textit{behavior–neurophysiology gap} in LM-based cognitive modeling: smaller LMs predict reading behavior better~\cite{Oh2023-zw}, while larger LMs excel in modeling neurophysiological data~\cite{Schrimpf2020-qa,michaelov2024revenge,Hosseini2024-jf}.
We suggest that this gap stems from the inconsistent treatment of internal layers (e.g., exclusive reliance on final layers, or inconsistent inclusion of intermediate layers).
If all the internal layers are focused on, even larger LMs have human-like responses through the lens of both behavioral and neurophysiological data.

Our exploration aligns with the common view that different human measures, operating on distinct timescales, reflect somewhat different stages of sentence processing.
For example, fast responses, such as first-pass gaze durations ($\sim$200ms), capture early-stage lexical processing~\cite{Calvo2002-nx}, while slower responses, such as N400 event-related potentials ($\sim$400ms), correspond to deeper semantic integration~\cite{Lau2008-pb,Kutas2011-dt,Nour-Eddine2024-ub}. 
Analogously, internal layers of LLMs may encode these temporal distinctions: earlier layers align with fast, shallow processes, while later layers correspond to slower, richer processes~\cite{Tenney2019-bb}.

In summary, our results suggest that larger LMs 
 provide superior cognitive plausibility in modeling both human behavior and neurophysiology data internally. 
In other words, shallower, cognitively plausible LMs are ``nested'' within LLMs. 
Broadly, these findings advocate the integration of cognitive modeling and mechanistic interpretability, encouraging a focus on layer-wise alignment with human measures.

\section{Related work}
\label{sec:background}

\subsection{Cognitive modeling and NLP}
\label{subsec:cog_model}

A key objective in linguistics is to understand how humans process language~\cite{Crocker2010-cp}, a goal that remains pertinent even in the era of LLMs.
According to perspectives outlined by \citet{marr1982}, information processing can be examined at three levels:
(i) the computational level: \textit{what is the goal of computation?}; 
(ii) the algorithmic level: \textit{how does the model achieve the goal?}; and 
(iii) the implementational level: \textit{how is it physically implemented?}. 
Humans can be viewed as an information processing model, and surprisal theory  --- humans continuously predict upcoming information during reading, with cognitive load incurred by unpredictable information ---~\cite{hale-2001-probabilistic,Levy2008Expectation-basedComprehension,Smith2013-ap} has accumulated its empirical evidence (\citealt{Smith2013-ap,FRANK20151,Shain2022-qv}; \textit{inter alia}).\footnote{Surprisal theory has also been critiqued~\cite{Van_Schijndel2021-sm,Huang2024-qe}, particularly for its failure to account for the cognitive load incurred in complex sentences. Our contribution is orthogonal to such criticism (see the Limitations section).} 
An orthogonal, algorithmic-level question regarding the surprisal theory is \textit{with what kind of algorithms and representations, humans predict upcoming information}. 
The NLP community has developed various methods to compute next-word probabilities, ranging from incremental parsers to LLMs, and researchers have tested their alignment with human reading data to provide insights into that question.
This alignment is typically investigated by analyzing which LMs compute surprisal $-\log p(\mathrm{word}|\mathrm{context})$ that correlates with human measures (e.g., word-by-word gaze durations) based on the surprisal theory. 

\subsection{Poorer fit of larger LMs' surprisal to human reading behavior}
\label{subsec:llm_cog}
In the days of much smaller LMs/parsers than modern LLMs (\citealt{hale-2001-probabilistic,Levy2008Expectation-basedComprehension,Smith2013-ap,frank2011insensitivity,Aurnhammer2019-fu,Merkx2020ComparingData}; \textit{inter alia.}), model-scaling generally improved their ability to simulate human sentence processing data~\cite{frank2011insensitivity,Goodkind2018PredictiveQuality,Wilcox2020OnBehavior,Wilcox2023-bb}. 
However, recent studies have questioned the generality of this scaling effect. 
The reversed trend was first found in typologically distant languages~\cite{kuribayashi-etal-2021-lower}, and even within English, further scaling up LMs have shown weaker alignment with human reading behavior~\cite{kuribayashi-etal-2022-context,Shain2022-qv,Oh2023-zw}. 
This \textit{bigger is not always better} phenomenon has become a key focus of LM-based cognitive modeling~\cite{Wilcox2024-qx}, with researchers investigating why LLMs appear cognitively implausible~\cite{kuribayashi-etal-2022-context,Oh2023-hj,Oh2023-zw,Oh2024-cc,nair2023words,kuribayashi-etal-2024-psychometric}.
In addition, from a more interdisciplinary view, mixed results are reported regarding such scaling effects.
For example, smaller LMs simulate reading behavior better~\cite{Oh2023-zw}, while larger LMs simulate neurophysiological data better~\cite{Schrimpf2020-qa,michaelov2024revenge,Hosseini2024-jf}. 
Opposite effect of instruction-tuning was also observed between brain data and behavioral data~\cite{aw2024instructiontuning,kuribayashi-etal-2024-psychometric}. 
We offer a perspective to address these negative scaling effects and behavior–neurophysiology gap, showing that the internal layers of larger LMs are more effective at modeling both behavioral and neurophysiological data.

\subsection{Human measure differences}
\label{subsec:layer_motivation}
We analyze the next-word predictions from the internal layers of LLMs in comparison with human sentence processing data. 
One motivation for this analysis is that different human measures, particularly at different time scales, may emphasize different stages of sentence processing~\cite{Witzel2012-nr,Lewis2005-hp,Vani2021-aj,Caucheteux2023-tx,McCurdy2024-ix}. 
LM internal layers, which are also computed sequentially, would be a natural counterpart to such multiple stages of processing~\cite{Tenney2019-bb}.
For example, eye movements reach the next word (or further) typically in $\sim$200ms before N400 brain signals peak at $\sim$400ms~\cite{Dimigen2011-gt}, suggesting that fast gaze durations may not reflect the cognitive load indexed by N400 signals~\cite{Rayner2009-aw}.

\section{Methods}

\subsection{Probabilities from internal layers}
\label{subsec:lens}
Our main proposal is to use the next-word probability $p(w_t|\bm w_{<t})$ of word $w_t$ in its context $\bm w_{<t}=[w_1,\cdots,w_{t-1}]^\top$ from internal layers in cognitive modeling, so we begin with how to extract internal surprisal.
We use two methods of \textit{logit-lens}~\cite{logitlens} and its sophisticated version of \textit{tuned-lens}~\cite{belrose2023eliciting}.
The first method, \textit{logit-lens}, extracts the probability of a word $w_t$ from a $d$-dimentaional internal representation $\bm h_{l,t}  \in \mathbb{R}^d$  at the $l$-th layer and time step $t$, as follows: 

{\small
\begin{align}
    \nonumber
    &p(w_t|\bm w_{<t}; \bm h_{l,t}) = \mathrm{LogitLens}(\bm h_{l,t})[\mathrm{id}(w_t)] \\
    &= \mathrm{softmax}(\bm W_U \mathrm{LayerNorm}(\bm h_{l,t}))[\mathrm{id}(w_t)]\mathrm{,}
    \label{eq:logit-lens}
\end{align}
}%
where $\bm W_U \in \mathbb{R}^{|\mathcal{V}| \times d}$ is an unembedding matrix obtained from LM's output layer, and $|\mathcal{V}| \in \mathbb{R}$ is model's vocabulary size. 
Simply put, the internal representaion $\bm h_{l,t}$ is mapped into output vocabulary space by applying $\bm W_U$ (i.e., skipping subsequent layers: $\bm h_{l+1, t}, \cdots, \bm h_{\mathrm{last}, t}$), and next-word probability is obtained in that space.
$\mathrm{LayerNorm}(\cdot):\mathbb{R}^d\to \mathbb{R}^d$ in Eq.~\ref{eq:logit-lens} is the layer normalization at the last layer, and [$\mathrm{id}(w_t)$] extracts the probability for $w_t$\footnote{
If a word is split into multiple subwords, accumulated surprisal is used. See Eq.2 in \citet{kuribayashi-etal-2021-lower}.} from the probability distribution over $\mathcal{V}$, obtained through the $\mathrm{softmax}(\cdot): \mathbb{R}^{|\mathcal{V}|}\to {[0,1]}^{|\mathcal{V}|}$ function.
The obtained probability (Eq.~\ref{eq:logit-lens}) is converted to surprisal $-\log p(w_t|\bm w_{<t}; \bm h_{l,t})$, and then used in the regression model to predict human reading data (\Cref{subsec:ppp}).

The second method, tuned-lens, extends logit-lens to handle the potential \textit{representational drifts} through layers. 
This technique introduces an additional linear transformation for each layer $l$ to mitigate the mismatch between the representation spaces of the $l$-th layer and the last layer:

{\small
\begin{align}
    \nonumber
    &p(w|\bm w_{<t}; \bm h_{l,t}) \\ &= \mathrm{LogitLens}(\bm W_l\bm h_{l,t}  + \bm b_l)[\mathrm{id}(w)]\mathrm{,}
    \label{eq:tuned}
\end{align}
}%
where $\bm W_l \in \mathbb{R}^{d\times d}$ and $\bm b_l \in \mathbb{R}^{d}$ are additionally trained to align the output of logit-lens with the last layer's next-word probability distribution on additional LM pretraining data.
We use publicly available tuned-lens parameters.
Notably, we do not fine-tune any part of the LMs for human data; instead, we observe the emerging correlations between next-word probabilities and human reading measures.

\subsection{Psychometric predictive power}
\label{subsec:ppp}

We evaluate the ability of surprisal values to predict word-by-word human cognitive responses, such as reading times or physiological signals (\Cref{subsec:data}). Following prior work (\citealt{wilcox-etal-2023-language,pimentel-etal-2022-effect}, \textit{inter alia}), this is done using linear regression models, motivated by surprisal theory~\cite{Smith2013-ap,Shain2022-qv}, which posits a linear relationship between surprisal and processing cost.

Formally, let $\bm w = [w_1, \dots, w_n]^\top$ denote the sequence of words in a dataset, and let $\bm y = [y_1, \dots, y_n]^\top \in \mathbb{R}^n$ be the corresponding word-by-word human measurements. We assess the predictive contribution of surprisal values $\bm s = [s(w_1), \dots, s(w_n)]^\top \in \mathbb{R}^n_{\ge 0}$, where surprisal is defined as $s(w_t) := -\log p(w_t|\bm w_{<t})$, using model probabilities as described in \Cref{subsec:lens}.

To evaluate the added benefit of surprisal over more primitive linguistic factors, we include a baseline feature vector $\bm b(w_t)$ for each word $w_t$:\footnote{Word length is measured in characters; word frequency is estimated using \texttt{word\_freq}~\cite{robyn_speer_2022_7199437}. We use a consistent set of baseline features across all datasets and models, with a few exceptions. For N400 data, a baseline amplitude term is added; for \citet{michaelov2023strong}'s EEG data, we additionally include electrode-level random effects.}

\begin{align}
\nonumber
\bm b(w_t) = [&\mathrm{length}(w_t), \mathrm{freq}(w_t), \mathrm{length}(w_{t-1}), \\
\nonumber
&\mathrm{freq}(w_{t-1}), \mathrm{length}(w_{t-2}), \mathrm{freq}(w_{t-2}), \\
& s(w_{t-1}), s(w_{t-2})]^\top.
\label{eq:baselines}
\end{align}

We include features of the two preceding words to account for spillover effects --- i.e., the processing difficulty of $w_{t-1}$ or $w_{t-2}$ can influence the response to $w_t$.

We train two nested linear regression models\footnote{Implemented using the \texttt{statsmodels} package~\cite{seabold2010statsmodels}. Some existing works include preceding words' surprisal values $s(w_{t-1}), s(w_{t-2})$ only in the full regression model (not in the reduced one; Eq.~\ref{eq:baselines}) when computing $\Delta$LL. We confirmed that this setting variation does not alter the conclusion, and at least in this paper, we adopt the setting to include these features in both full and reduced regression models.}: (i) a full model including both surprisal and baseline features; and (ii) a reduced model using only the baseline features. The coefficients are estimated via ordinary least squares. Model fit is quantified by the log-likelihood under a Gaussian noise assumption, and the difference in log-likelihoods between the two models --- denoted $\Delta\mathrm{LL}$ --- reflects the isolated contribution of surprisal.

A higher $\Delta\mathrm{LL}$ indicates greater predictive power, and we refer to this value as the \textit{psychometric predictive power} (PPP). The central question of our study is: which LM layer yields surprisal values with the strongest PPP?

\begin{figure*}[t]
    \centering
    \includegraphics[width=1.0\linewidth]{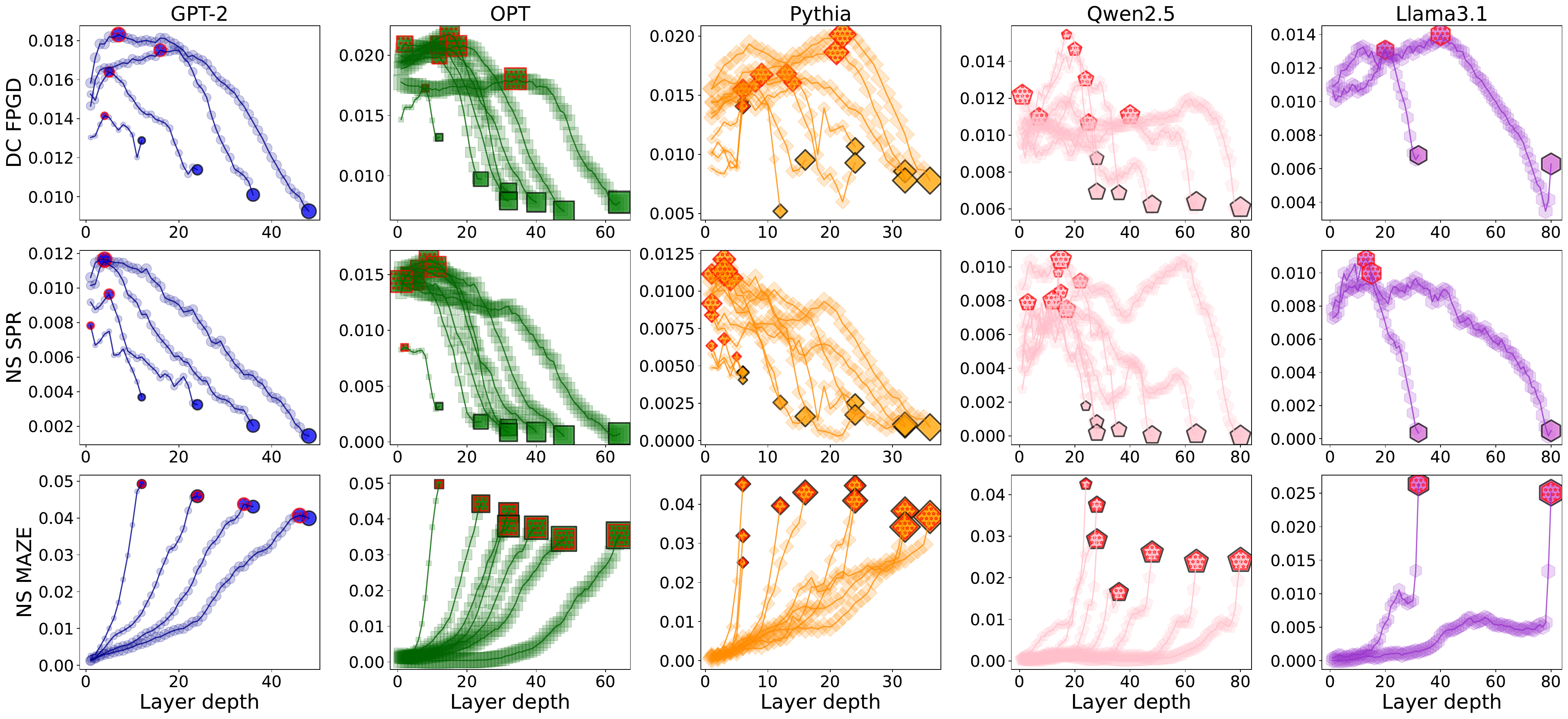}
    \caption{Relationships between layer depth (x-axis) and $\Delta$LL (y-axis) for each LM in three datasets: FPGD in DC, SPR in NS, and MAZE in NS. The graphs are separated by model families and data. Each line corresponds to a different model, and one with larger markers and more layers corresponds to the bigger model within each model family. For each model, the best layer is indicated with a red edge line and shading patterns, and the last layer is indicated with a black edge.  The graph starts at the first layer, not at the embedding layer. The results are based on logit-lens.}
    \label{fig:reading_time}
\end{figure*}

\begin{table*}[t]
    \centering
    \scriptsize
    \tabcolsep0.07cm
\begin{tabular}{llrrrrrp{0.2cm}rrrrr}
\toprule
 &  & \multicolumn{5}{c}{PPP\_logit-lens} & & \multicolumn{5}{c}{PPP\_tuned-lens} \\
  \cmidrule(lr){3-7}
  \cmidrule(lr){9-13}
Stimuli & Measure & 0-0.2 & 0.2-0.4 & 0.4-0.6 & 0.6-0.8 & 0.8-1 & & 0-0.2 & 0.2-0.4 & 0.4-0.6 & 0.6-0.8 & 0.8-1 \\
  \midrule
\multirow{1}{*}{DC} & FPGD~\cite{kennedy2003dundee} & 14.13 & 14.78 & \textbf{14.92} & 13.38 & 9.84 & & \textbf{17.10} & 16.32 & 15.39 & 13.53 & 10.49 \\
\cmidrule(r){1-2} \cmidrule(lr){3-7} \cmidrule(lr){9-13}
\multirow{2}{*}{NS} & SPR~\cite{Futrell2021-wr} & \textbf{9.85} & 9.75 & 8.44 & 5.68 & 2.67 & & \textbf{8.93} & 7.03 & 5.11 & 3.44 & 2.33 \\
 & MAZE~\cite{Boyce2023AmazeON} & 1.18 & 3.00 & 5.69 & 12.06 & \textbf{23.77} & & 9.70 & 17.56 & 24.15 & 32.86 & \textbf{39.63} \\
 \cmidrule(r){1-2} \cmidrule(lr){3-7} \cmidrule(lr){9-13}
\multirow{2}{*}{ZuCO} & FPGD~\cite{Hollenstein2018-rm} & 38.10 & \textbf{38.13} & 35.59 & 29.82 & 15.94 & & \textbf{30.48} & 27.16 & 22.56 & 17.29 & 8.77 \\
 & N400~\cite{Hollenstein2018-rm} & 0.07 & 0.12 & 0.15 & \textbf{0.18} & 0.16 & & 0.20 & 0.32 & \textbf{0.34} & 0.29 & 0.18 \\
\cmidrule(r){1-2} \cmidrule(lr){3-7} \cmidrule(lr){9-13}
\multirow{3}{*}{UCL} & SPR~\cite{frank2013reading} & \textbf{22.88} & 22.21 & 19.30 & 11.45 & 4.77 & & \textbf{15.78} & 8.92 & 4.87 & 2.53 & 1.27 \\
 & FPGD~\cite{frank2013reading} & 22.11 & \textbf{23.39} & 22.83 & 15.77 & 6.53 & & \textbf{16.28} & 14.48 & 11.87 & 9.47 & 5.57 \\
 & N400~\cite{FRANK20151} & \textbf{56.86} & 38.04 & 22.77 & 16.07 & 22.58 & & 11.31 & 6.12 & 16.19 & 29.49 & \textbf{37.11} \\
\cmidrule(r){1-2} \cmidrule(lr){3-7} \cmidrule(lr){9-13}
\multirow{3}{*}{Fillers} & SPR~\cite{Vasishth2010-ji} & 7.83 & 10.89 & \textbf{14.39} & 14.18 & 14.35 & & 8.60 & 10.47 & 11.36 & 11.86 & \textbf{13.33} \\
 & FPGD~\cite{Vasishth2010-ji} & 6.66 & 5.83 & 6.48 & 7.31 & \textbf{10.36} & & 8.94 & 10.91 & 12.91 & 13.81 & \textbf{14.00} \\
 & MAZE~\cite{Hahn2022-ib} & 5.39 & 3.01 & 5.08 & 21.97 & \textbf{60.89} & & 9.96 & 28.27 & 52.00 & 73.38 & \textbf{88.64} \\
\cmidrule(r){1-2} \cmidrule(lr){3-7} \cmidrule(lr){9-13}
\multirow{1}{*}{Michaelov+,2024} & N400~\cite{michaelov2023strong} & 0.88 & 1.42 & \textbf{1.91} & 1.68 & 0.91 & & 0.95 & 1.51 & \textbf{1.70} & 1.38 & 0.99 \\
\cmidrule(r){1-2} \cmidrule(lr){3-7} \cmidrule(lr){9-13}
\multirow{1}{*}{Federmeier+,2007} & N400~\cite{Federmeier2007-qg} & 0.77 & 3.11 & 8.59 & 18.05 & \textbf{25.80} & & 1.49 & 5.22 & 13.06 & 24.48 & \textbf{28.71} \\
\multirow{1}{*}{W\&F,2012} & N400~\cite{Wlotko2012-eh} & \textbf{0.35} & 0.19 & 0.10 & 0.09 & 0.12 & & \textbf{0.51} & 0.27 & 0.12 & 0.05 & 0.11 \\
\multirow{1}{*}{Hubbard+,2019} & N400~\cite{Hubbard2019-vz} & 0.18 & 0.23 & 0.23 & \textbf{0.25} & 0.17 & & 0.11 & 0.12 & 0.22 & \textbf{0.36} & 0.33 \\
\multirow{1}{*}{S\&F,2022} & N400~\cite{Szewczyk2022-ds} & 0.11 & 0.15 & 0.38 & 0.90 & \textbf{1.40} & & 0.16 & 0.33 & 0.77 & 1.29 & \textbf{1.42} \\
\multirow{1}{*}{Szewczyk+,2022} & N400~\cite{Szewczyk2022-cd} & 1.21 & 2.91 & 4.43 & 6.40 & \textbf{8.04} & & 2.12 & 3.58 & 5.52 & 8.10 & \textbf{8.93} \\
\bottomrule
\end{tabular}
        \caption{All the results. The $\Delta$LL scores are averaged by the layer relative depth, e.g., first 20\% of layers as ``0-0.2,'' across models, and the best relative layer range for each data is highlighted in bold. $\Delta$LLs are multiplied by 1000 for brevity.} 
        \label{tbl:results}
\end{table*}

\section{Experimental settings}
\label{sec:settings}

\subsection{Human data}
\label{subsec:data}
We use 15 human reading datasets, listed in Table~\ref{tbl:results},  (we additionally use MECO in~\Cref{subsec:cross-lingual}), which include human measurements from various methods: self-paced reading time (SPR), first-pass gaze duration (FPGD), Maze task processing time (MAZE), and electroencephalography (EEG; specifically the N400 component). 
The datasets share a common format: each word $w_t$ is annotated with $\mathrm{Cost}(w_t) \in \mathbb{R}$ representing the human cognitive load associated with it. 
Our corpus selection aligns with recent studies~\cite{kuribayashi-etal-2024-psychometric,michaelov2024revenge,de-Varda2024-yh,McCurdy2024-ix}.\footnote{We applied the same preprocessing as~\citet{kuribayashi-etal-2024-psychometric} (DC, NS), \citet{de-Varda2024-yh} (UCL), \citet{Hahn2022-ib} (Fillers), and \citet{michaelov2024revenge} (N400). For ZuCO, we only used the naturalistic reading part, and for its N400, we averaged the values at the central electrode between 300-500ms during the first pass over a word.}

SPR is measured by presenting sentences through a sliding word-by-word window, with participants pressing a button to advance. 
FPGD, a key eye-tracking measure, represents the total time from first fixating on a word to moving to another word. 
Maze processing time is measured during a task requiring participants to select the plausible continuation of a sentence, offering a controlled alternative to naturalistic reading. 
EEG measures brain activity, with N400 reflecting the negative brain potential peaking around 400ms after word presentation. 
These are the common measures employed to study expectation-based sentence processing.
SPR, FPGD, and MAZE are categorized as human \textit{behavioral} data, while EEG falls under \textit{neurophysiological} data.

To minimize confounding factors between stimulus data and human measures, we included datasets with multi-layered annotations across multiple human measures. 
These include the Natural Stories Corpus~\cite{Futrell2021-wr} with SPR\footnote{We use the version (2025-05-12) without the misalignment problem (see \url{https://github.com/languageMIT/naturalstories}).} and MAZE data~\cite{Boyce2023AmazeON}, ZuCO corpus~\cite{Hollenstein2018-rm} with FPGD and N400 data, UCL Corpus~\cite{frank2013reading} annotated with SPR, FPGD, and N400 data~\cite{FRANK20151}, and filler sentences from~\citet{Hahn2022-ib} annotated with SPR, FPGD, and MAZE data~\cite{Vasishth2010-ji,Hahn2022-ib}. In particular, the FPGD and N400 data in ZuCO were simultaneously recorded from the same human subjects, which likely minimized confounding factors.
 As is common in preprocessing, we exclude data points with zero SPR/FPGD/MAZE value.
Human data for each token in the corpus were averaged across subjects prior to analysis, following recent practices \cite{pimentel-etal-2022-effect,Oh2023-zw,kuribayashi-etal-2024-psychometric,de-Varda2024-yh}.

\subsection{Language models}
\label{subsec:model}
We evaluate 30 open-source LMs including bilion-scale ones: GPT-2 (124M, 355M, 774M, and 1.5B parameters)~\cite{Radford_undated-nn}, OPT (125M, 1.3B, 2.7B, 6.7B, 13B, 30B, and 66B parameters)~\cite{opt}, Pythia (14M, 31M, 70M, 160M, 410M, 1B, 1.4B, 2.8B, 6.9B, and 12B parameters)~\cite{biderman2023pythia}, Qwen2.5 (0.5B, 1.5B, 3B, 7B, 14B, 32B, 72B), and Llama-3.1 (8B and 70B).
See Appendix~\ref{app:lms} for details.
For tuned-lens experiments (\Cref{subsec:lens}), we use 14 of these models based on the availability of pre-trained tuned lenses.\footnote{We used parameters released in \url{https://huggingface.co/spaces/AlignmentResearch/tuned-lens/tree/main}. Specifically, we used GPT-2 124M, 774M, 1.5B; OPT 125M, 1.3B, 6.7B; and Pythia 70M, 160M, 410M, 1B, 1.4B, 2.8B, 6.9B, 12B, in the tuned-lens experiments. Notably, their actual implementation used the representation of $\bm h_{l,t}\bm W'_l + \bm h_{l,t} + \bm b_l=\bm h_{l,t}(\bm W'_l + \bm 1) + \bm b_l$, but we omit the identity matrix $\bm 1 \in \mathbb{R}^{d\times d}$ in Eq.~\ref{eq:tuned} by overriding $\bm W_l=\bm W'_l + \bm 1$.}
The number of internal layers in our models ranges from 6 to 80. Results for surprisal from the embedding layer are excluded as these generally show a bad fit with human data. %

\begin{figure*}[t]
    \centering
    \begin{subfigure}[t]{0.49\textwidth}
    \centering
        \includegraphics[width=1.0\linewidth]{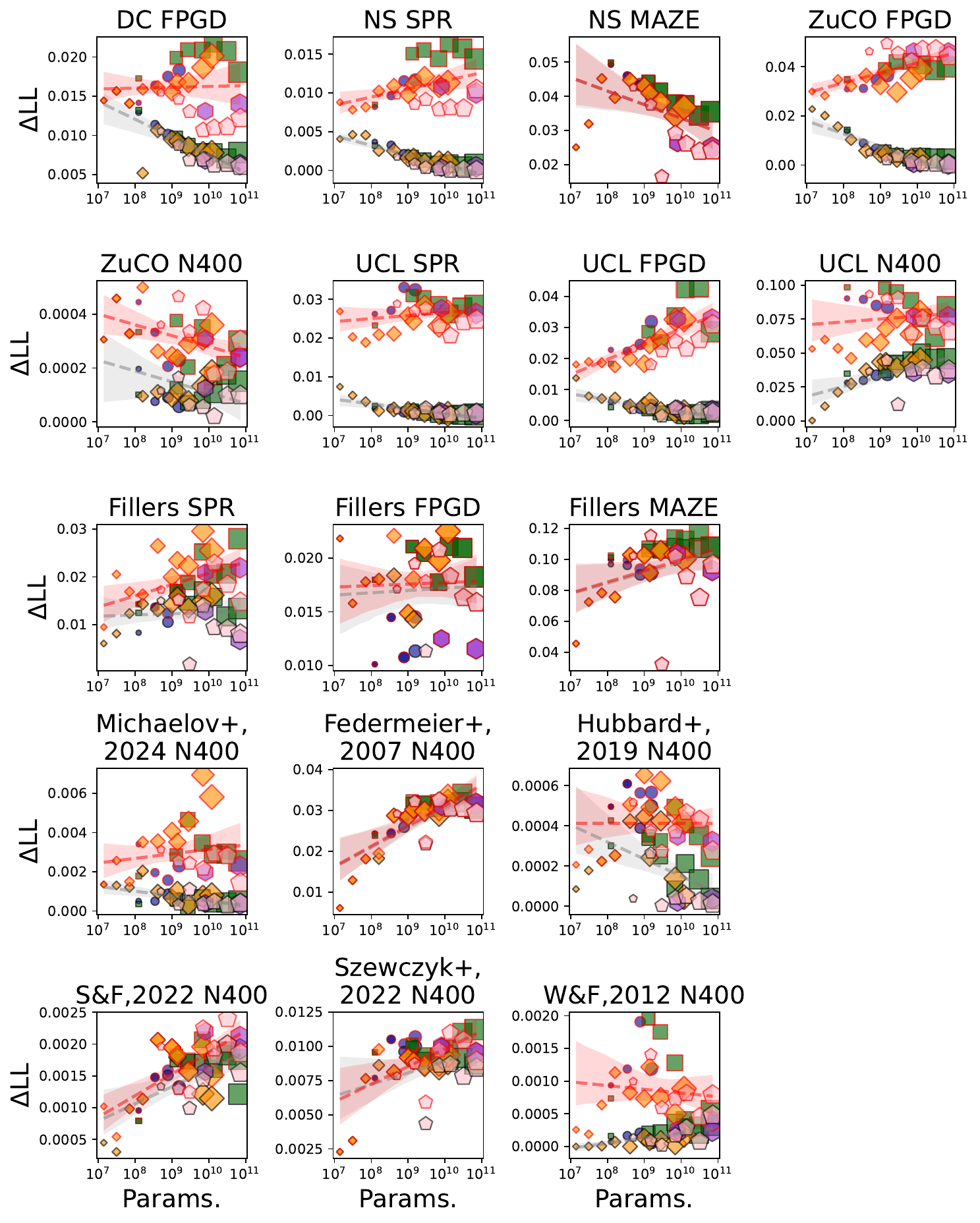}
        \caption{Logit-lens}
    \end{subfigure}
    \hfill
    \begin{subfigure}[t]{0.49\textwidth}
    \centering
        \includegraphics[width=1.0\linewidth]{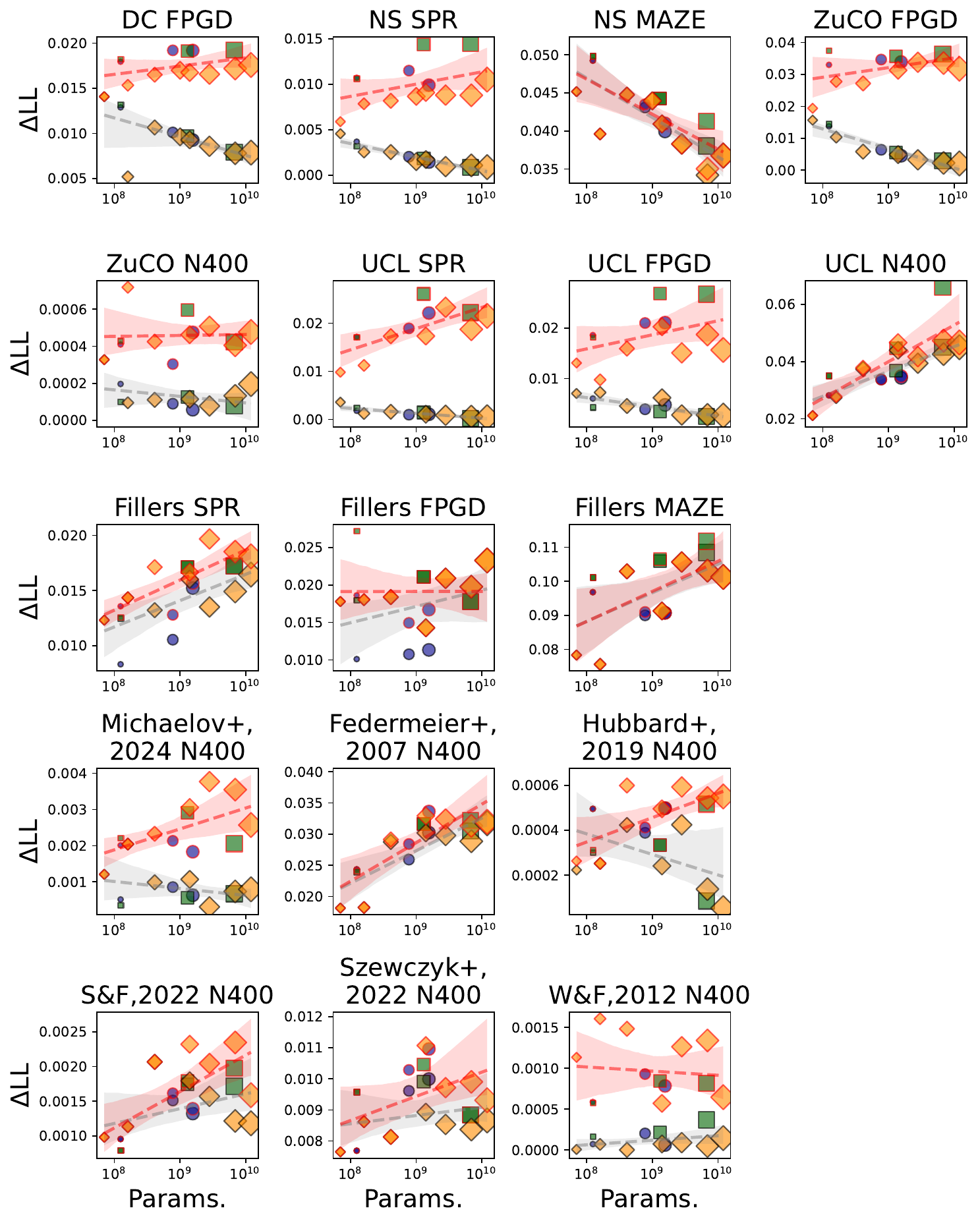}
        \caption{Tuned-lens}
    \end{subfigure}
    \caption{
Scaling effect between model size (parameter counts in log scale) and $\Delta$LL. Each marker corresponds to each LM's $\Delta$LL score from its best layer (red edge) or the last layer (black edge).
The regression lines show the scaling effects, and the red line is for best-layer's $\Delta$LL while the grey one is for the last layer's. 
The maker type (shape/size/color) follows Figure~\ref{fig:overview}. 
    }
    \label{fig:params_ppp}
\end{figure*}

\section{Experimental results}
\label{sec:results}

Figures~\ref{fig:reading_time} (\Cref{subsec:best_layer}) and~\ref{fig:params_ppp} (\Cref{subsec:params_ppp}) summarize our main findings, with comprehensive results in Table~\ref{tbl:results}.

\subsection{Last layer does not yield the best $\Delta$LL}
\label{subsec:best_layer}

We first revisit the experimental settings (DC and NS datasets) of \citet{Oh2023-zw}. 
The top two line graphs in Figure~\ref{fig:reading_time} depict the layer-wise $\Delta$LL for each LM. 
A consistent pattern emerges in FPGD and SPR data: the $\Delta$LL decreases toward the final layer (the rightmost, less-transparent, large markers), indicating that the last layer often yields the lowest score compared to the internal layers of the same model. 
These results challenge the assumption, widely adopted in existing studies, that the last layer is the most reliable indicator of an LM's cognitive plausibility. 

At the same time, changing the human measurement, i.e., from SPR to MAZE in the NS data (the second and third rows in Figure~\ref{fig:reading_time}), flips the trend.
This indicates that the aligned layers can easily change when using different human measurement methods, and suggests that different aspects of human sentence processing correlate with different LM layers, depending on the human measurement method.
Such measurement--layer interaction effect is further looked into in~\Cref{subsec:layer_and_measure}.

Table~\ref{tbl:results} presents a detailed breakdown of all the datasets, averaging $\Delta$LL scores across relative layer positions (e.g., 0--0.2 for the first 20\% of layers). 
Table~\ref{tbl:results} includes the results with logit-lens and tuned-lens, which yielded generally consistent patterns.
As shown in Figure~\ref{fig:reading_time}, the best-performing layer is often not the final one (0.8--1.0), and the optimal layer varies based on the type of measurement and stimuli. 
For instance, SPR and FPGD data are best modeled by earlier layers, whereas MAZE processing times and N400 signals are better captured by later layers (as motivated in \Cref{subsec:layer_motivation}). 
Furthermore, there are also stimulus-dependent biases --- the optimal layer for the UCL dataset is among the earlier layers, while for the Fillers dataset, it lies among the later layers (perhaps associated with the specific complexity of the stimulus).

\subsection{Revisiting LM-scaling effects in cognitive modeling with internal layers}
\label{subsec:params_ppp}

We revisit the question with our extended focus on model internals: what kind of LMs yield the best $\Delta$LL from their internals?  As the field is particularly interested in the relationship with model scaling~\cite{Goodkind2018PredictiveQuality,Oh2023-zw}, we examine the relationship between LM parameter size (x-axis) and $\Delta$LL (y-axis) for two scenarios:  (1) using the last layer's $\Delta$LL (grey lines), reproducing previous findings; and (2) considering the best $\Delta$LL layer identified in this study (red lines).

Figure~\ref{fig:params_ppp} illustrates these two relationships. The grey lines align with prior findings relying on the last layer~\cite{Oh2023-zw,michaelov2024revenge}, showing mixed scaling effects, where larger LMs do not consistently outperform smaller ones. However, the red lines reveal a positive scaling trend: larger LMs achieve equal or better $\Delta$LL compared to smaller LMs when internal layers are taken into account. 
The Pearson correlation coefficients between parameter numbers and $\Delta$LL from the best layers were significantly larger than zero on average, across settings.\footnote{We collected the correlation coefficients between logarithmic number of model parameters and $\Delta$LL from the best layers from 34 settings of $\{\mathrm{dataset}\}\times\{\mathrm{lens}\}$, and one-sample t-test shows that these coefficients are, on average, significantly larger than zero ($\text{p-value}<0.05$).}
This suggests that when the analysis extends to internal layers, the $\Delta$LL ranking flips, revealing that larger LMs are seemingly more cognitively plausible. In other words, larger LMs embed cognitively plausible, smaller LMs within their internal. One notable exception is the MAZE processing time in the NS dataset~\cite{Boyce2023AmazeON}, where a strictly negative scaling effect persists, even when internal layers are considered. 
PPL–$\Delta$LL relationships\footnote{Perplexity (PPL), a general quality measure of LMs, is a geometric mean of next-word probabilities over data $L$: $\prod_{t=1}^{|L|} p(w_t|\bm w_{<t})^{1/|L|}$. The PPL--$\Delta$LL relationship has long been investigated~\cite{frank2011insensitivity,Goodkind2018PredictiveQuality,kuribayashi-etal-2021-lower,Oh2023-zw}.} are additionally shown in Figure~\ref{fig:ppl_ppp} in the Appendix, which also show that the poor $\Delta$LL of larger, more accurate LMs is mitigated.

\begin{table*}[t]
    \centering
    \scriptsize
    \tabcolsep0.05cm
\begin{tabular}{lrrrrrrrrrrrrrrrrrr}
\toprule
  & GPT2 & \multicolumn{6}{c}{OPT}& \multicolumn{5}{c}{Pythia} & \multicolumn{4}{c}{Qwen} & \multicolumn{2}{c}{Llama-3} \\
  \cmidrule(lr){2-2}
  \cmidrule(lr){3-8}
  \cmidrule(lr){9-13}
  \cmidrule(lr){14-17}
  \cmidrule(lr){18-19}
 Data  & XL & 1.3B & 2.7B & 6.7B & 13B & 30B & 66B & 1B & 1.4B & 2.8B & 6.9B & 12B & 3B & 7B & 32B & 72B & 8B & 70B \\
  \midrule
 DC FPGD~\cite{kennedy2003dundee} & 0.80 & 0.80 & 0.82 & 0.76 & 0.73 & 0.76 & 0.78 & 0.00 & 0.32 & 0.36 & 0.73 & 0.81 & 0.10 & 0.26 & 0.72 & 0.94 & 0.89 & 0.73 \\
NS SPR~\cite{Futrell2021-wr} & 0.82 & 0.80 & 0.82 & 0.76 & 0.76 & 0.78 & 0.82 & 0.65 & 0.60 & 0.64 & 0.73 & 0.95 & 0.93 & 0.95 & 0.93 & 0.97 & 0.98 & 0.73 \\
ZuCO FPGD~\cite{Hollenstein2018-rm} & 0.80 & 0.84 & 0.88 & 0.79 & 0.73 & 0.78 & 0.86 & 0.65 & 0.60 & 0.55 & 0.76 & 0.97 & 0.97 & 0.97 & 0.98 & 0.97 & 0.98 & 0.78 \\
UCL SPR~\cite{frank2013reading} & 0.78 & 0.80 & 0.79 & 0.76 & 0.76 & 0.78 & 0.80 & 0.71 & 0.52 & 0.64 & 0.70 & 0.97 & 0.97 & 0.97 & 0.98 & 0.97 & 0.99 & 0.70 \\
UCL FPGD~\cite{frank2013reading} & 0.94 & 0.88 & 0.82 & 0.79 & 0.83 & 0.88 & 0.83 & 0.59 & 0.56 & 0.58 & 0.70 & 0.97 & 0.93 & 0.97 & 0.94 & 0.97 & 0.99 & 0.73 \\
\bottomrule
\end{tabular}
    \caption{How likely $\Delta$LL from internal layers outperformed the previous best $\Delta$LL (achieved within the same model family, relying on their last layers). The results are focused on billion-scale models and behavioral data with somewhat drastic flips in LM-scaling effects for $\Delta$LLs. }
    \label{tab:win_rate}
\end{table*}

\section{Analyses}
\label{sec:analysis}
We conduct several follow-up analyses to support and clarify the findings reported in~\Cref{sec:results}.

\subsection{How easily can good layers be found?}
\label{subsec:layer_selection}
One immediate concern in~\Cref{subsec:params_ppp} is how many numbers of internal layers yield a good $\Delta$LL; perhaps, we just observed outliers as best-$\Delta$LL values in Figure~\ref{fig:params_ppp}.
Given this concern, we analyze the number of internal layers that outperform the previously best $\Delta$LL achieved by the last layer within the same model family. 
Table~\ref{tab:win_rate} presents the win rate of internal layers' $\Delta$LL against the respective previous best score.
The win rate is typically around 80\%, indicating a significant proportion of internal layers achieved good $\Delta$LL scores. 
These findings support that the cognitive plausibility of LLMs has been underestimated and that our argument (Figure~\ref{fig:params_ppp}) was not based on specific outlier layers but reflected a broader trend across many internal layers.

\subsection{Layer depth and human measures}
\label{subsec:layer_and_measure}
We observed systematic tendencies in the relationship between layer depth and human measurement methods. 
For instance, FPGD aligns better with earlier layers, whereas N400 aligns better with later layers, as summarized in Table~\ref{tbl:results}. 
To statistically validate this relationship, we trained a linear regression model to explain $\Delta$LL scores from our 23,154 experimental settings $\{\mathrm{dataset}\}\times \{\mathrm{model}\} \times \{\mathrm{layer}\}$ with the following features for each setting $s$: \{$\mathrm{stimuli}(s)$, $\mathrm{model}(s)$, $\mathrm{lens}(s)$, $\mathrm{layer\_depth}(s)$, $\mathrm{measure}(s)$, $\mathrm{layer\_depth}(s)$:$\mathrm{measure}(s)$\}.
Here, $\mathrm{stimuli}$ represents the source stimuli of the data (``Stimuli'' column in Table~\ref{tbl:results}), $\mathrm{model}$ encodes the model name, $\mathrm{layer\_depth}$ is the depth of the layer where the $\Delta$LL is obtained, $\mathrm{measure}$ encodes the human measurement method (``Measure'' column in Table~\ref{tbl:results}), and $\mathrm{lens}$ indicates whether the logit-lens or tuned-lens was used. The term $\mathrm{layer\_depth} \times \mathrm{measure}$ captures the interaction between effective layer depth and human measures, which is of interest. Note that $\mathrm{measure}$ is a categorical variable, with $\mathrm{SPR}$ serving as the dummy category.

\begin{figure}[t]
    \centering
    \begin{subfigure}[t]{0.23\textwidth}
        \centering
        \includegraphics[width=\textwidth]{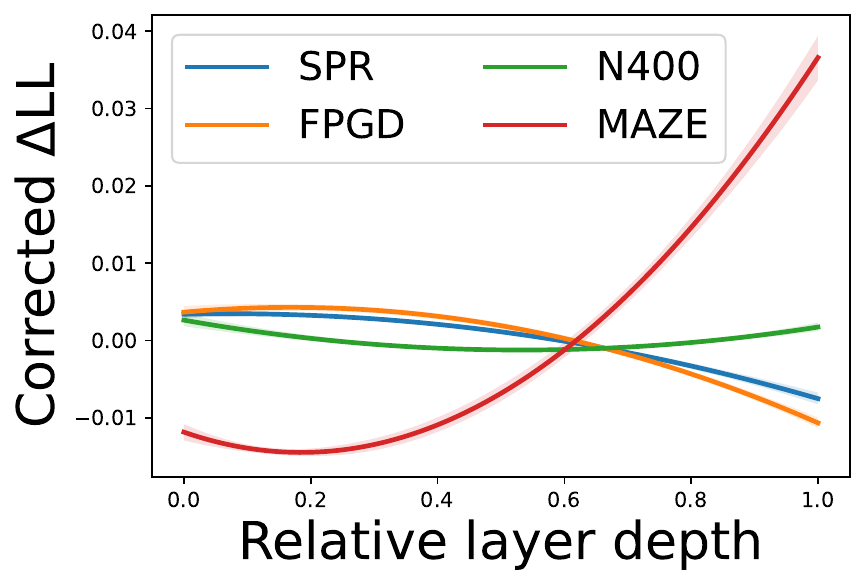}
        \caption{All tokens.}
        \label{fig:diff}
    \end{subfigure}
    \begin{subfigure}[t]{0.23\textwidth}
        \centering
        \includegraphics[width=\textwidth]{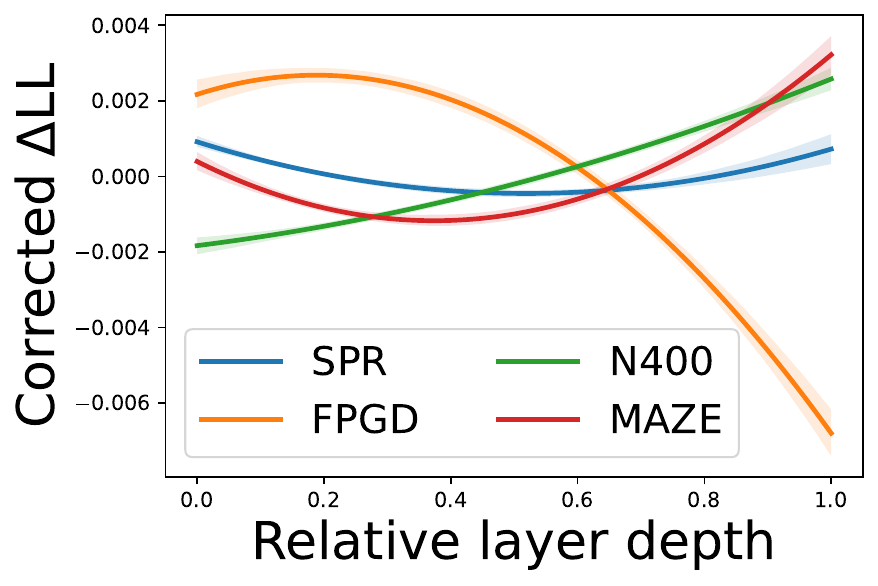}
        \caption{Clause-final tokens.}
        \label{fig:diff_last_token}
    \end{subfigure}
    \caption{Relationship between $\Delta$LL and relative layer depth for each human measure. Different measures are associated with different layers; for example, good $\Delta$LLs for FPGD are achieved in earlier layers, while those for MAZE are in the latter layers.}
    \label{fig:layer_diff}
\end{figure}

The coefficients for $\mathrm{layer\_depth} \times \mathrm{N400}$ and $\mathrm{layer\_depth} \times  \mathrm{MAZE}$ were significantly larger than zero ($\text{p-value}<0.001$), while that for $\mathrm{layer\_depth} \times \mathrm{FPGD}$ is significantly lower than $\mathrm{layer\_depth} \times \mathrm{SPR}$ (see full regression results in  Table~\ref{tbl:coefficient_interaction} in the Appendix).
This confirms that $\mathrm{SPR}$ aligns with earlier layers than those yielding a good fit with $\mathrm{FPGD}$, and $\mathrm{N400}$ and $\mathrm{MAZE}$ align with later layers than $\mathrm{FPGD}$. 
We also visualize the relationships between $\Delta$LL and relative layer depth for each human measure in Figure~\ref{fig:diff} (polynomial fit using 2nd-order term). 
Here, we use corrected $\Delta$LLs that are computed by subtracting variances explained by factors other than $\mathrm{measure}$ based on the regression model. 
The lines also indicate the differences across different human measures, e.g., good $\Delta$LL for MAZE is clearly associated with the latter layers.

\begin{table*}[t]
    \centering
    \footnotesize
    \tabcolsep=0.07cm
\begin{tabular}{llrrrrrp{0.2cm}rrrrr}
\toprule
 &  & \multicolumn{5}{c}{Logit-lens ($\Delta$LL)} && \multicolumn{5}{c}{Tuned-lens ($\Delta$LL)} \\
Stimuli & Measure & 0-0.2 & 0.2-0.4 & 0.4-0.6 & 0.6-0.8 & 0.8-1.0 && 0-0.2 & 0.2-0.4 & 0.4-0.6 & 0.6-0.8 & 0.8-1.0 \\
 \cmidrule(r){1-1} \cmidrule(r){2-2}  \cmidrule(lr){3-3} \cmidrule(lr){4-4}  \cmidrule(lr){5-5} \cmidrule(lr){6-6}  \cmidrule(lr){7-7}   \cmidrule(lr){9-9}  \cmidrule(lr){10-10} \cmidrule(lr){11-11} \cmidrule(lr){12-12}  \cmidrule(lr){13-13}  
\multirow{1}{*}{DC} & FPGD~\cite{kennedy2003dundee} & 1.24 & 1.37 & \textbf{1.42} & 1.31 & 1.00 & & 1.59 & 1.60 & \textbf{1.65} & 1.58 & 1.30 \\
\cmidrule(r){1-2} \cmidrule(lr){3-7} \cmidrule(lr){9-13}
\multirow{2}{*}{NS} & SPR~\cite{Futrell2021-wr} & \textbf{0.45} & 0.38 & 0.33 & 0.29 & 0.21 & & 0.16 & 0.12 & 0.12 & 0.18 & \textbf{0.25} \\
 & MAZE~\cite{Boyce2023AmazeON} & 0.35 & 0.36 & 0.42 & 0.62 & \textbf{0.84} & & 0.67 & 0.70 & 0.76 & 1.02 & \textbf{1.30} \\
\cmidrule(r){1-2} \cmidrule(lr){3-7} \cmidrule(lr){9-13}
\multirow{2}{*}{ZuCO} & FPGD~\cite{Hollenstein2018-rm} & 38.10 & \textbf{38.13} & 35.59 & 29.82 & 15.94 & & \textbf{30.48} & 27.16 & 22.56 & 17.29 & 8.77 \\
 & N400~\cite{Hollenstein2018-rm} & 0.07 & 0.12 & 0.15 & \textbf{0.18} & 0.16 & & 0.20 & 0.32 & \textbf{0.34} & 0.29 & 0.18 \\
\cmidrule(r){1-2} \cmidrule(lr){3-7} \cmidrule(lr){9-13}
\multirow{3}{*}{UCL} & SPR~\cite{frank2013reading} & \textbf{3.09} & 2.53 & 2.12 & 1.21 & 0.50 & & \textbf{1.76} & 0.89 & 0.50 & 0.27 & 0.15 \\
 & FPGD~\cite{frank2013reading} & 7.51 & 7.89 & \textbf{8.33} & 6.67 & 3.56 & & 4.85 & \textbf{4.86} & 4.60 & 4.26 & 2.66 \\
 & N400~\cite{FRANK20151} & 3.58 & 1.66 & 1.99 & 5.20 & \textbf{11.95} & & 1.12 & 3.85 & 8.31 & 12.75 & \textbf{18.17} \\
\cmidrule(r){1-2} \cmidrule(lr){3-7} \cmidrule(lr){9-13}
\multirow{3}{*}{Fillers} & SPR~\cite{Vasishth2010-ji} & 0.28 & 0.33 & 0.62 & 1.86 & \textbf{5.78} & & 0.88 & 2.19 & 3.32 & 5.83 & \textbf{10.31} \\
 & FPGD~\cite{Vasishth2010-ji} & 0.30 & 0.25 & 0.26 & 0.59 & \textbf{1.29} & & 0.28 & 0.42 & 0.52 & 1.40 & \textbf{2.04} \\
 & MAZE~\cite{Hahn2022-ib} & 3.54 & 2.41 & 1.51 & 3.83 & \textbf{8.99} & & 1.15 & 1.94 & 5.52 & 10.64 & \textbf{12.53} \\
 \bottomrule
\end{tabular}
\caption{Results for the same settings as Table~\ref{tbl:results}, except that only sentence/clause-final tokens are targeted. Some N400 data are omitted because they initially targeted only sentence/clause-final tokens. $\Delta$LLs are multiplied by 1000 for brevity.} 
\label{tbl:results_clause_final}
\end{table*}

\subsection{Are results biased by targeted tokens?}
\label{subsec:clause_final}

A potential confound in~\Cref{subsec:layer_and_measure} stems from differences in targeted tokens for different human measures. For instance, N400 data are typically recorded only at sentence-final tokens to preprocess continuous, time-series EEG data, potentially introducing biases specific to sentence-final tokens, such as wrap-up effects~\cite{Just1980AComprehension,Rayner2000TheReading,meister-etal-2022-analyzing}. In contrast, behavioral measures are recorded across tokens within a sentence. To address this potential confound, we conducted additional experiments targeting sentence/clause-final tokens for all measures, even for SPR, FPGD, and MAZE. Sentence/clause-final tokens are ideitified using a constituency parser~\cite{kitaev-etal-2019-multilingual,kitaev-klein-2018-constituency} and punctuations (e.g., ``.'', ``,''), following~\citet{meister-etal-2022-analyzing}.

Table~\ref{tbl:results_clause_final} presents the results. Even when analyses were restricted to sentence-final tokens, earlier layers continued to align better with SPR and FPGD data. Regression analysis (same as \Cref{subsec:layer_and_measure}) confirmed that the coefficients for $\mathrm{layer\_depth} \times \mathrm{N400}$ and $\mathrm{layer\_depth} \times \mathrm{MAZE}$ remained significantly larger than those for $\mathrm{layer\_depth} \times \mathrm{SPR}$, and  the coefficient for $\mathrm{layer\_depth} \times \mathrm{FPGD}$ was lower than that for $\mathrm{layer\_depth} \times \mathrm{SPR}$ (see full regression results in Table~\ref{tbl:coefficient_interaction_last_token} in the Appendix). These patterns are visualized in Figure~\ref{fig:diff_last_token}.

\subsection{When are earlier layers advantageous?}
\label{subsec:context}

We further explore more general trends on when and why earlier layers' suprisal aligns better with human reading data.
Following~\citet{Oh2023-zw}, we analyze by-token squared residual errors from regression models predicting human data (\Cref{subsec:ppp}). 
We specifically use the largest data of DC (FPGD) and identify tokens where the use of the best internal layer, rather than the last layer, notably reduces errors.\footnote{\citet{Oh2023-zw} reported a misalignment between MSEs (residual errors) and log-likelihood scores due to the Euclidean norm penalty adopted in the \texttt{lme4} package. This did not arise in our analysis with \texttt{statsmodels}.} 
To address this question, we fit a linear regression model to explain the decreases in squared residual errors observed in each data point $w_t$ with an LM $\theta$, with the following by-token linguistic properties as features: \{$\mathrm{model}(\theta)$, $\mathrm{length}(w_t)$, $\mathrm{freq}(w_t)$, $\mathrm{position}(w_t)$, $\mathrm{POS}(w_t)$\}.

Results show that decreases in modeling error are associated with less frequent, longer words (see Table~\ref{tab:error_regression} in Appendix for full regression results). 
This aligns with prior observations~\cite{Oh2024-cc} that LLMs tend to predict infrequent tokens with overly confident surprisals, and surprisal from internal layers mitigates this issue.

\begin{figure}[t]
    \centering
    \includegraphics[width=1.0\linewidth]{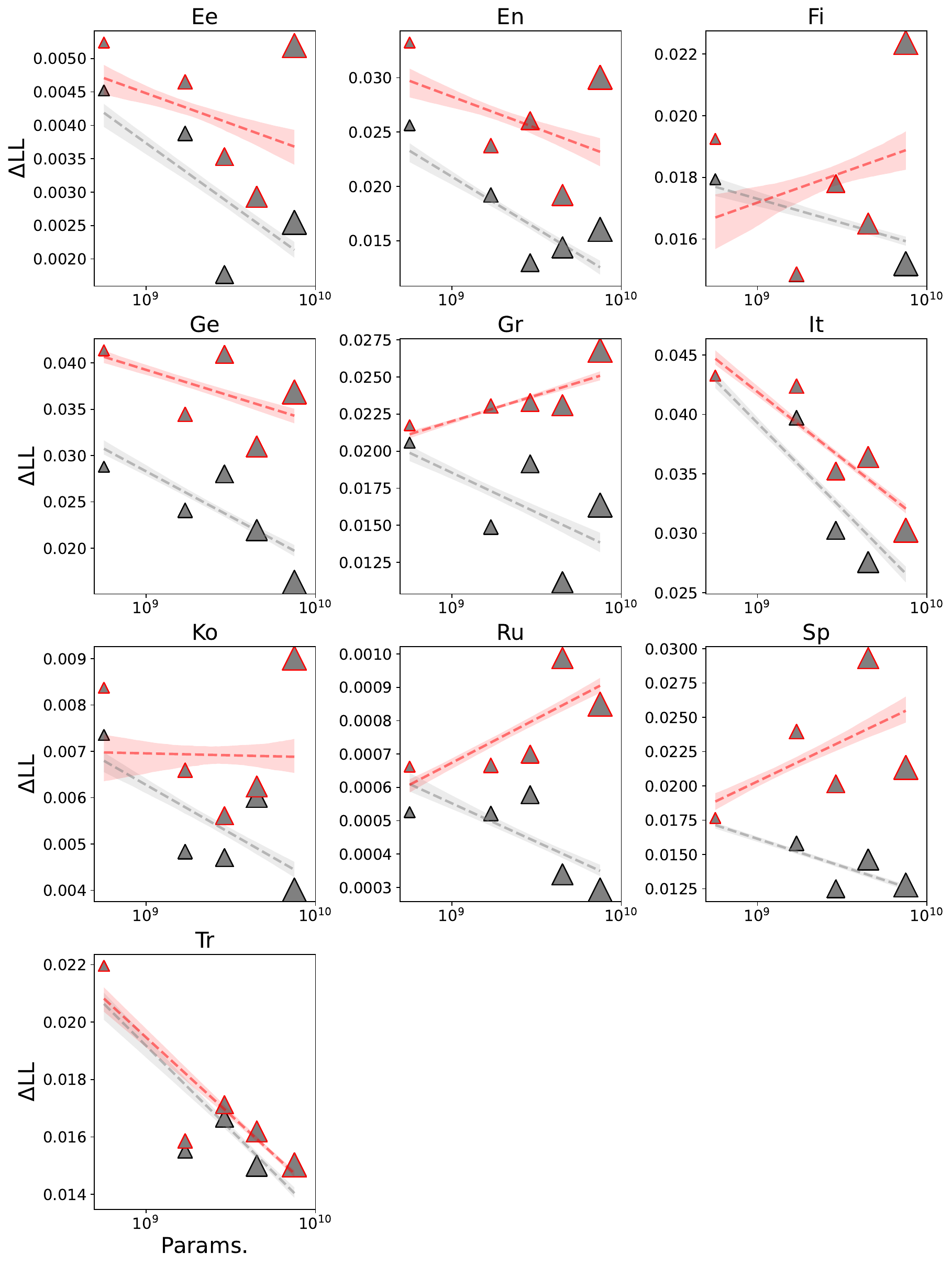}
    \caption{
Scaling effect between $\Delta$LL and parameter counts in MECO. The grey lines are results relying on the last layer's $\Delta$LLs, and the red lines rely on the best internal layers' $\Delta$LLs.
In all the languages, the negative correlation between parameter size and $\Delta$LL is mitigated to some extent, and in four languages, the relationship turned out to be positive.
    }
    \label{fig:params_ppp_meco}
\end{figure}

\begin{figure}[t]
    \centering
    \begin{subfigure}[t]{0.23\textwidth}
    \centering
        \includegraphics[width=\linewidth]{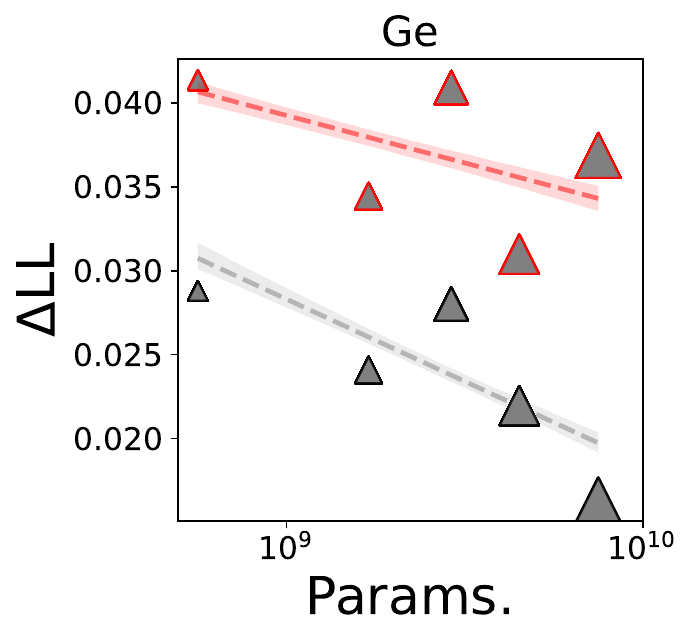}
        \caption{Multilingual models}
    \end{subfigure}
    \hfill
    \begin{subfigure}[t]{0.23\textwidth}
        \centering
        \includegraphics[width=\linewidth]{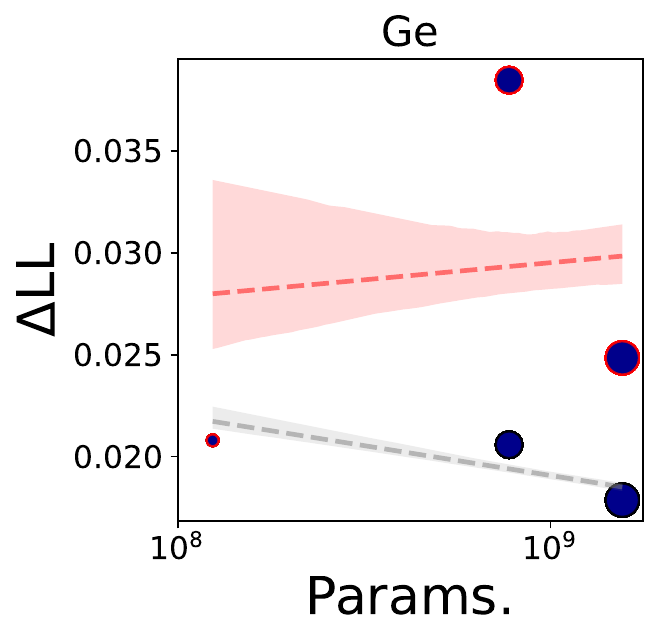}
        \caption{Monolingual models}
    \end{subfigure}
    \caption{Scaling effects in MECO German part}
    \label{fig:ge-results}
\end{figure}

\subsection{Mutlilingual generallity}
\label{subsec:cross-lingual}
Our main experiments were limited to the English language --- what about the advantage of internal surprisal in other languages?
Following the experiments of \citet{de-varda-marelli-2023-scaling}, we also explored the multilingual generality of our results, using FPGD data in ten languages recorded in MECO~\cite{siegelman2022expanding}
Five variants of multilingual XGLMs~\cite{lin-etal-2022-shot} are used.\footnote{We excluded Dutch, Hebrew, and Norwegian because these are not supported by XGLMs. In addition, due to the unavailability of tuned-lens for XGLMs, only logit-lens was used.}
Figure~\ref{fig:params_ppp_meco} shows the relationship between parameter size and $\Delta$LL, where the red markers and lines are based on the best internal layers, and the gray ones are from the final layers.
First, the best layer typically outperforms the last layer of the respective model (86\%=43/50 of the settings), as the best $\Delta$LLs (red markers) are generally higher than the last layer's $\Delta$LLs (grey markers) in Figure~\ref{fig:params_ppp_meco}, reproducing the findings in~\Cref{subsec:best_layer}.

Second, the negative correlation score between parameter size and $\Delta$LL consistently increases in all the languages, which partially aligns with~\Cref{subsec:params_ppp}.
Nevertheless, the scaling effect is still negative in some languages.
We suspect that these mixed results might be biased by multilingual LMs, which are reported to process every language within the English subspace in their middle layers~\cite{Wendler2024-wr}, leading to biased probability estimates. 
As a case study to handle this concern, we compared the results of multilingual XGMs (left part of Figure~\ref{fig:ge-results}) with those of German monolingual GPT-2s (right part of Figure~\ref{fig:ge-results}; see Appendix~\ref{app:lms} for model details).
The negative scaling effects are flipped to be positive with the use of monolingual LMs, implying that the inclusion of monolingual LMs further enhances the advantage of the internal layer's surprisal.

\section{Discussion}
We lastly discuss the implications of our findings, connections to existing studies, and future works.

\subsection{Connection to context sensitivity}
\label{subsec:working_memory}
We propose a perspective linking our findings to the context sensitivity of humans and LMs during sentence processing. 
Earlier layers may better model human reading behavior because they are less contextualized, reflecting the human-like tendency to process sentences under working memory constraints.

\paragraph{Working memory limitations in humans}
Human sentence processing is constrained by limited cognitive resources, relying on selective and efficient use of context~\cite{Lewis2005-hp,Lieder2019-xo,Futrell2020Lossy-ContextProcessing,Hahn2022-ib}. 
Recent studies indicate that the MAZE task imposes greater working memory demands, requiring more extensive context use than SPR or FPGD~\cite{Hahn2022-ib,McCurdy2024-ix}. 
This aligns with our observation that MAZE processing times are better modeled by later, more context-sensitive layers (see the next paragraph), whereas SPR and FPGD align with earlier, less contextualized layers.

\begin{figure}[t]
    \centering
    \includegraphics[width=0.9\linewidth]{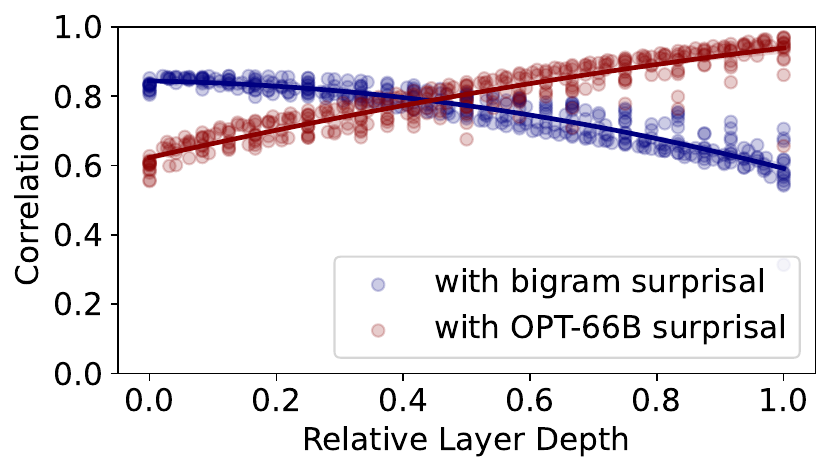}
    \caption{
    The markers correspond to all the internal layers of our targeted LMs, which are sorted by relative layer depth (x-axis).  Two types of scores (y-axis) are plotted: (i) Pearson correlation coefficient between each layer's surprisal vs.\ less-contextualized bigram surprisal (blue); and (ii) each layer's surprisal vs.\ well-contextualized LLM surprisal (red). We used tuned-lens results.
    }
    \label{fig:contextualization}
\end{figure}

\paragraph{Working memory limitations in LMs}
Modern neural LMs are not optimized to conserve cognitive resources and often rely excessively on context information, resulting in superhuman predictions~\cite{kuribayashi-etal-2022-context,Oh2024-cc}. 
However, internal layers may exhibit a human-like moderation of context use. 
The NLP community has observed that LMs gradually enhance contextualization across layers, from shallow representations in early layers to deeply contextualized representations in later layers (\citealt{Brunner2019-so,Ethayarajh2019-zy,Toneva2019-ul}).

We also confirmed the by-layer context-sensitivity of surprisal by analyzing the correlation between (i) intermediate layers' surprisal and less-contextualized bigram surprisal\footnote{Bigram LM is trained on OpenWebText~\cite{Gokaslan2019OpenWeb} with the KenLM toolkit~\cite{Heafield2011-ce}}; and (ii) intermediate layers' surprisal and more contextualized surprisal from the LLM with the lowest PPL (OPT-66B). 
Figure~\ref{fig:contextualization} shows the above two types of correlations for each model's layer, with the x-axis as the relative layer depth.
This shows that earlier layers correlate more strongly with bigram surprisal (Pearson correlation coefficient $r$ for this relationship is $-$0.92), while later layers align with more accurate, well-contextualized surprisal ($r$=0.95).
These findings reinforce the idea that earlier layers exhibit limited context sensitivity, while later layers are more contextualized and better suited for modeling data like MAZE, which demands higher contextualization.

\subsection{Connection to spill-over effects}
There is another possible explanation about the alignment of earlier/later layers with SPR/MAZE that was discussed in~\Cref{subsec:context}.
Self-paced reading and eye-tracking measures often exhibit \textit{spillover effects}, where the processing of one word influences subsequent words~\cite{Rayner1998-rw}. 
This suggests that the comprehension of a word extends beyond the immediate moment, and the associated reading times may only capture an early stage of processing.
In contrast, the MAZE task~\cite{Forster2009-dp,Boyce2023AmazeON}, which measures the time taken to select a plausible continuation from two candidates, mitigates spillover effects and is thus expected to reflect the full process of word processing. 
Our findings --- early LM layers are more closely aligned with gaze durations and self-paced reading times, while later layers show a stronger alignment with MAZE --- align with this perspective. 

\subsection{Connection to LM-brain alignment study}
Brain imaging data (e.g., fMRI) and reading behavior data have complementary advantages.
Generally speaking, the former has a high spatial resolution (in which part of the brain particular processing is performed), while the latter has a high temporal resolution (how long does the processing take).
In the former context of fMRI modeling research, layer-wise LM-human alignments, similarly to us, have been attempted~\cite{Toneva2019-ul,Schrimpf2020-qa,Caucheteux2023-tx} and suggested that different brain areas better align with different LM layers.
Our study is orthogonal to these studies with more focus on the temporal alignment between reading-time and layer-time scales, which is concurrently explored in~\citet{hu2025signatures}.

One additional difference with the above-mentioned fMRI studies is that they typically trained linear regression models to predict brain activity directly using $d$-dimentional LM internal representations $\bm h \in \mathbb{R}^d$ as features, instead of using surprisal measures $-\log p(\mathrm{word}|\mathrm{context})\in \mathbb{R}_{\ge 0}$.
Thus, their results are not directly comparable with existing surprisal-based studies and may rather suffer from a confounding factor of different $d$ for different LMs when precisely discussing LM-scaling effects.

\subsection{Connection to early exits of neural models' prediction}
\label{subsec:adaptive}
In the engineering context, predictions from internal layers of deep neural networks (i.e., early exits of the results from internal layers) are used to improve inference efficiency by avoiding their overthinking~\cite{Graves2016-dp,Banino2021-oq}, which is also called adaptive computation time~\cite{icml19shallowdeepnetworks,Zhou2020-vz}. Such a technique has also recently been employed to enhance interpretability research to identify at which layer a particular prediction shapes~\cite{logitlens,dar-etal-2023-analyzing,belrose2023eliciting}.

\subsection{Limitations toward surprisal theory}
Recent studies have raised several issues, orthogonal to our study, on surprisal-based cognitive modeling, for example, on tokenizations~\cite{nair2023words,Giulianelli2024-ms,Oh2024-cf}, LM training scenario~\cite{Oh2023-hj}, as well as more refined indicators of word predictability~\cite{pimentel-etal-2022-effect,Giulianelli2024-wd,Opedal2024-si,Meister2024-vq}.
More critically, the reliance solely on surprisal obtained from LMs tends to underestimate the significant slowdown of sentence processing against syntactic amguity and grammatical violation~\cite{Wilcox2021-gy,Van_Schijndel2021-sm,Huang2024-qe,Wang2024-po}.
This study, as an initial foray, is focused on cognitive modeling on naturalistic reading corpora, but including such experiments with controlled materials will enrich our findings on internal surprisals.

Despite such concerns regarding surprisal theory, we would like to highlight that our core proposal --- using probability from internal layers in cognitive modeling --- is not limited to surprisal theory in human sentence processing modeling, but can contribute to any cognitive modeling work that uses probability-related measures and neural models.
For example, our analysis can easily be extended to other metrics such as entropy, entropy reduction~\cite{hale2016information}, or any new probability-based measures combining probabilities from different layers~\cite{hu2025signatures}.
Furthermore, our idea to focus on internal layers may also be combined with linguistically-motivated neural models, such as deep neural incremental parsers~\cite{Dyer2016RecurrentGrammars,Yoshida2021-rc}, to potentially combine linguistic theories (e.g., theory of syntax) with surprisal theory.
Thus, any limitations of surprisal theory itself do not undermine the core contributions of this work.

\section{Conclusions}
Recent cognitive modeling studies have demonstrated a worse fit of surprisal from larger LMs to human reading time.
In this paper, we argue that these negative results stem from the exclusive focus on the \textit{final layers} of LMs.
Instead, we observe that those from \textit{internal layers} comparably or even better fit with human behavior and neurophysiology data, suggesting that the cognitive plausibility of larger LMs has been underestimated.
Furthermore, different human measurements align with different layers, implying an intriguing parallel between temporal dynamics in human sentence processing and LM internal layers.

\section*{Limitations}
The experiments can be extended to more types of human measures, such as first fixation time, go-past reading time, eye regressions, total fixation time, EEG data other than the N400 potential~\cite{Federmeier2007-qg} as well as FMRI data~\cite{Shain2021-yk}.
As a first step, we have begun with the measures of SPR, FPGD, N400, and MAZE, as they are typically used in cognitive modeling.
We also conducted an exploratory analysis with other human measures (e.g., P600, second-pass gaze duration) on the UCL corpus (Table~\ref{tbl:results_other_variables} in Appendix). 
For such an extended scope, we cannot confirm the implied relationship between fast–slow human responses and early–late LM layers, and our findings appear to hold primarily for commonly-analyzed human measures of SPR, FPGD, N400, and MAZE.
It is also a critical question in psycholinguistics whether the distinction between ``fast'' (gaze durations) and ``slow'' (N400) human measures adopted in this study is truly related to some cognitive costs or the time indeed humans took to process the word.
Especially for EEG, the delay may perhaps be just due to technical properties regarding the temporal resolutions.

\section*{Acknowledgment}
We appreciate our TACL action editor and reviewers. We also thank Ryo Yoshida for constructive feedback on the earlier version of this work.
This work was supported by JSPS KAKENHI Grant Number JP24H00087; JST PRESTO Grant Number JPMJPR21C2; JSPS Grant-in-Aid for Early-Career Scientists Grant Number JP23K16938.

\bibliography{custom}
\bibliographystyle{acl_natbib}

\clearpage
\appendix

\onecolumn

\section{Details on psychometric predictive power}
\label{app:ppp}

Here, we explain the $\Delta$LL (PPP) score more formally. Recall that we quantify the predictive contribution of word-by-word surprisal to human cognitive responses using a log-likelihood-based score, $\Delta$LL. Let $\bm{w} = [w_1, \dots, w_n]^\top$ denote a sequence of words, and let $\bm{y} = [y_1, \dots, y_n]^\top$ be the corresponding word-by-word human cognitive costs, e.g., reading times.

For each word $w_i$, we compute a surprisal value $s(w_i) \in \mathbb{R}$, and a vector of baseline linguistic features $b(w_i) \in \mathbb{R}^d$. We consider two linear models: a \textit{full model} using both surprisal and baseline features, and a \textit{reduced model} using only baseline features.

In both cases, we estimate the coefficients using ordinary least squares (OLS). The full model solves:
\[
\hat{\phi}_s, \hat{\bm{\phi}}_b = \arg\min_{\phi_s, \bm{\phi}_b} \sum_{i=1}^n \left( y_i - \phi_s s(w_i) - \bm{\phi}_b^\top b(w_i) \right)^2.
\]

We assume homoscedastic Gaussian noise and evaluate the total log-likelihood of the fitted model using the maximum likelihood estimate of the variance:
\[
\text{LL}_{\text{full}} = -\frac{n}{2} \log(2\pi \hat{\sigma}_{\text{full}}^2) - \frac{n}{2}, \quad \text{where} \quad \hat{\sigma}_{\text{full}}^2 = \frac{1}{n} \sum_{i=1}^n \left( y_i - \hat{\phi}_s s(w_i) - \hat{\bm{\phi}}_b^\top b(w_i) \right)^2.
\]

The reduced model is obtained by regressing $y$ on baseline features alone:
\[
\hat{\bm{\phi}}_b' = \arg\min_{\bm{\phi}_b} \sum_{i=1}^n \left( y_i - \bm{\phi}_b^\top b(w_i) \right)^2,
\]
with corresponding log-likelihood:
\[
\text{LL}_{\text{baseline}} = -\frac{n}{2} \log(2\pi \hat{\sigma}_{\text{baseline}}^2) - \frac{n}{2}, \quad \text{where} \quad \hat{\sigma}_{\text{baseline}}^2 = \frac{1}{n} \sum_{i=1}^n \left( y_i - \hat{\bm{\phi}}_b'^\top b(w_i) \right)^2.
\]

We define the \textit{psychometric predictive power} of surprisal as the difference in log-likelihood between the full and reduced models:
\[
\Delta \text{LL} = \text{LL}_{\text{full}} - \text{LL}_{\text{baseline}} = \frac{n}{2} \log \left( \frac{\hat{\sigma}_{\text{baseline}}^2}{\hat{\sigma}_{\text{full}}^2} \right).
\]

This $\Delta$LL value quantifies the contribution of surprisal to predicting human sentence processing data, with higher values indicating stronger predictive power.

\section{Models}
\label{app:lms}
Table~\ref{tbl:lms} shows the exact models we used.

\section{Details on cross-lingual experiments}
\label{app:cross_lingual}

We target FPGD data from MECO in 13 languages~\cite{siegelman2022expanding} using five multilingual XGLMs~\cite{lin-etal-2022-shot} (564M, 1.7B, 2.9B, 4.5B, 7.5B).
For the German part, we used monolingual German models in Table~\ref{tbl:lms}.
We analyze the $\Delta$LL of surprisal from their internal layers. 
Due to the unavailability of tuned-lenses for XGLMs, only logit-lenses were used.
Figure~\ref{fig:params_ppp_meco} shows the relationship between paramter numbers and $\Delta$LL in each language.

\begin{table}[ht]
    \centering
    \footnotesize
    \begin{tabular}{lll}
    \toprule
       Model & URL & \#params \\
       \cmidrule(lr){1-1} \cmidrule(lr){2-2} \cmidrule(lr){3-3}
       GPT2 & \url{https://huggingface.co/gpt2} & 117M \\
       GPT2‑medium & \url{https://huggingface.co/gpt2-medium} & 345M \\
       GPT2‑large & \url{https://huggingface.co/gpt2-large} & 774M \\
       GPT2‑xl & \url{https://huggingface.co/gpt2-xl} & 1B \\
       \cmidrule(lr){1-3}
       OPT‑125m & \url{https://huggingface.co/facebook/opt-125m} & 125M \\
       OPT‑1.3b & \url{https://huggingface.co/facebook/opt-1.3b} & 1.3B \\
       OPT‑2.7b & \url{https://huggingface.co/facebook/opt-2.7b} & 2.7B \\
       OPT‑6.7b & \url{https://huggingface.co/facebook/opt-6.7b} & 6.7B \\
       OPT‑13b & \url{https://huggingface.co/facebook/opt-13b} & 13B \\
       OPT‑30b & \url{https://huggingface.co/facebook/opt-30b} & 30B \\
       OPT‑66b & \url{https://huggingface.co/facebook/opt-66b} & 66B \\
       \cmidrule(lr){1-3}
       Pythia‑14m‑deduped & \url{https://huggingface.co/EleutherAI/pythia-14m-deduped} & 14M \\
       Pythia‑31m‑deduped & \url{https://huggingface.co/EleutherAI/pythia-31m-deduped} & 31M \\
       Pythia‑70m‑deduped & \url{https://huggingface.co/EleutherAI/pythia-70m-deduped} & 70M \\
       Pythia‑160m‑deduped & \url{https://huggingface.co/EleutherAI/pythia-160m-deduped} & 160M \\
       Pythia‑410m‑deduped & \url{https://huggingface.co/EleutherAI/pythia-410m-deduped} & 410M \\
       Pythia‑1b‑deduped & \url{https://huggingface.co/EleutherAI/pythia-1b-deduped} & 1B \\
       Pythia‑1.4b‑deduped & \url{https://huggingface.co/EleutherAI/pythia-1.4b-deduped} & 1.4B \\
       Pythia‑2.8b‑deduped & \url{https://huggingface.co/EleutherAI/pythia-2.8b-deduped} & 2.8B \\
       Pythia‑6.9b‑deduped & \url{https://huggingface.co/EleutherAI/pythia-6.9b-deduped} & 6.9B \\
       Pythia‑12b‑deduped & \url{https://huggingface.co/EleutherAI/pythia-12b-deduped} & 12B \\
       \cmidrule(lr){1-3}
       Llama‑3.1‑8B & \url{https://huggingface.co/meta-llama/Llama-3.1-8B-Instruct} & 8B \\
       Llama‑3.1‑70B & \url{https://huggingface.co/meta-llama/Llama-3.1-70B-Instruct} & 70B \\
       \cmidrule(lr){1-3}
       Qwen2.5‑0.5B & \url{https://huggingface.co/Qwen/Qwen2.5-0.5B} & 500M \\
       Qwen2.5‑1.5B & \url{https://huggingface.co/Qwen/Qwen2.5-1.5B} & 1.5B \\
       Qwen2.5‑3B & \url{https://huggingface.co/Qwen/Qwen2.5-3B} & 3B \\
       Qwen2.5‑7B & \url{https://huggingface.co/Qwen/Qwen2.5-7B} & 7B \\
       Qwen2.5‑14B & \url{https://huggingface.co/Qwen/Qwen2.5-14B} & 14B \\
       Qwen2.5‑32B & \url{https://huggingface.co/Qwen/Qwen2.5-32B} & 32B \\
       Qwen2.5‑72B & \url{https://huggingface.co/Qwen/Qwen2.5-72B} & 72B \\       
        \cmidrule(lr){1-3}
       XGLM-564M & \url{https://huggingface.co/facebook/xglm-564M} & 564M \\
       XGLM-1.7B & \url{https://huggingface.co/facebook/xglm-1.7B} & 1.7B \\
       XGLM-2.9B & \url{https://huggingface.co/facebook/xglm-2.9B} & 2.9B \\
       XGLM-4.5B & \url{https://huggingface.co/facebook/xglm-4.5B} & 4.5B \\
       XGLM-7.5B & \url{https://huggingface.co/facebook/xglm-7.5B} & 7.5B \\
       \cmidrule(lr){1-3}
        \multirow{3}{*}{German LMs} &  \url{https://huggingface.co/benjamin/gerpt2} & 124M \\
         & \url{https://huggingface.co/benjamin/gerpt2-large} & 774M \\
         & \url{https://huggingface.co/malteos/gpt2-xl-wechsel-german} & 1.5B \\
        \bottomrule
    \end{tabular}
    \caption{LM details}
    \label{tbl:lms}
\end{table}

\clearpage

\twocolumn

\begin{table}[ht]
    \centering
    \tiny
    \tabcolsep=0.02cm
\renewcommand{\arraystretch}{0.7}
\begin{subtable}[h]{0.49\textwidth}
\centering
\scalebox{0.9}{
\begin{tabular}{lcccccc}
\toprule
                                           & \textbf{coef} & \textbf{std err} & \textbf{t} & \textbf{P$> |$t$|$} & \textbf{[0.025} & \textbf{0.975]}  \\
\midrule
\textbf{Intercept}                         &       0.0125  &        0.001     &    20.725  &         0.000        &        0.011    &        0.014     \\
\textbf{stimuli[T.Federmeier+,
2007]}      &       0.0040  &        0.001     &     7.520  &         0.000        &        0.003    &        0.005     \\
\textbf{stimuli[T.Fillers]}                &       0.0002  &        0.000     &     0.570  &         0.569        &       -0.001    &        0.001     \\
\textbf{stimuli[T.Hubbard+,
2019]}         &      -0.0079  &        0.001     &   -14.755  &         0.000        &       -0.009    &       -0.007     \\
\textbf{stimuli[T.Michaelov+,
2024]}       &      -0.0068  &        0.001     &   -12.666  &         0.000        &       -0.008    &       -0.006     \\
\textbf{stimuli[T.NS]}                     &      -0.0069  &        0.000     &   -14.314  &         0.000        &       -0.008    &       -0.006     \\
\textbf{stimuli[T.S\&F,2022]}              &      -0.0075  &        0.001     &   -13.955  &         0.000        &       -0.009    &       -0.006     \\
\textbf{stimuli[T.Szewczyk+,
2022]}        &      -0.0033  &        0.001     &    -6.076  &         0.000        &       -0.004    &       -0.002     \\
\textbf{stimuli[T.UCL]}                    &       0.0090  &        0.000     &    22.701  &         0.000        &        0.008    &        0.010     \\
\textbf{stimuli[T.W\&F,2012]}              &      -0.0079  &        0.001     &   -14.816  &         0.000        &       -0.009    &       -0.007     \\
\textbf{stimuli[T.ZuCO]}                   &       0.0036  &        0.000     &     8.735  &         0.000        &        0.003    &        0.004     \\
\textbf{model[T.Llama-3.1-8B]}             &       0.0010  &        0.001     &     1.707  &         0.088        &       -0.000    &        0.002     \\
\textbf{model[T.Qwen2.5-0.5B]}             &      -0.0004  &        0.001     &    -0.689  &         0.491        &       -0.002    &        0.001     \\
\textbf{model[T.Qwen2.5-1.5B]}             &      -0.0007  &        0.001     &    -1.101  &         0.271        &       -0.002    &        0.001     \\
\textbf{model[T.Qwen2.5-14B]}              &      -0.0004  &        0.001     &    -0.822  &         0.411        &       -0.001    &        0.001     \\
\textbf{model[T.Qwen2.5-32B]}              &       0.0002  &        0.000     &     0.473  &         0.637        &       -0.001    &        0.001     \\
\textbf{model[T.Qwen2.5-3B]}               &      -0.0012  &        0.001     &    -2.129  &         0.033        &       -0.002    &    -9.41e-05     \\
\textbf{model[T.Qwen2.5-72B]}              &       0.0003  &        0.000     &     0.731  &         0.465        &       -0.001    &        0.001     \\
\textbf{model[T.Qwen2.5-7B]}               &      -0.0003  &        0.001     &    -0.481  &         0.630        &       -0.001    &        0.001     \\
\textbf{model[T.gpt2]}                     &       0.0010  &        0.001     &     1.568  &         0.117        &       -0.000    &        0.002     \\
\textbf{model[T.gpt2-large]}               &       0.0020  &        0.000     &     4.266  &         0.000        &        0.001    &        0.003     \\
\textbf{model[T.gpt2-medium]}              &       0.0008  &        0.001     &     1.242  &         0.214        &       -0.000    &        0.002     \\
\textbf{model[T.gpt2-xl]}                  &       0.0023  &        0.000     &     5.367  &         0.000        &        0.001    &        0.003     \\
\textbf{model[T.opt-1.3b]}                 &       0.0026  &        0.001     &     5.084  &         0.000        &        0.002    &        0.004     \\
\textbf{model[T.opt-125m]}                 &       0.0024  &        0.001     &     3.778  &         0.000        &        0.001    &        0.004     \\
\textbf{model[T.opt-13b]}                  &       0.0028  &        0.001     &     5.301  &         0.000        &        0.002    &        0.004     \\
\textbf{model[T.opt-2.7b]}                 &       0.0028  &        0.001     &     4.902  &         0.000        &        0.002    &        0.004     \\
\textbf{model[T.opt-30b]}                  &       0.0023  &        0.001     &     4.513  &         0.000        &        0.001    &        0.003     \\
\textbf{model[T.opt-6.7b]}                 &       0.0027  &        0.000     &     5.639  &         0.000        &        0.002    &        0.004     \\
\textbf{model[T.opt-66b]}                  &       0.0027  &        0.000     &     5.721  &         0.000        &        0.002    &        0.004     \\
\textbf{model[T.pythia-1.4b-deduped]}      &       0.0010  &        0.001     &     1.992  &         0.046        &     1.65e-05    &        0.002     \\
\textbf{model[T.pythia-12b-deduped]}       &       0.0023  &        0.000     &     4.920  &         0.000        &        0.001    &        0.003     \\
\textbf{model[T.pythia-14m]}               &      -0.0017  &        0.001     &    -1.508  &         0.132        &       -0.004    &        0.000     \\
\textbf{model[T.pythia-160m-deduped]}      &      -0.0015  &        0.001     &    -2.352  &         0.019        &       -0.003    &       -0.000     \\
\textbf{model[T.pythia-1b-deduped]}        &       0.0010  &        0.001     &     1.367  &         0.172        &       -0.000    &        0.002     \\
\textbf{model[T.pythia-1b-deduped-v0]}     &       0.0040  &        0.002     &     2.415  &         0.016        &        0.001    &        0.007     \\
\textbf{model[T.pythia-2.8b-deduped]}      &       0.0016  &        0.000     &     3.349  &         0.001        &        0.001    &        0.003     \\
\textbf{model[T.pythia-31m]}               &      -0.0015  &        0.001     &    -1.391  &         0.164        &       -0.004    &        0.001     \\
\textbf{model[T.pythia-410m-deduped]}      &       0.0008  &        0.001     &     1.476  &         0.140        &       -0.000    &        0.002     \\
\textbf{model[T.pythia-6.9b-deduped]}      &       0.0018  &        0.000     &     3.826  &         0.000        &        0.001    &        0.003     \\
\textbf{model[T.pythia-70m-deduped]}       &      -0.0016  &        0.001     &    -1.911  &         0.056        &       -0.003    &     4.05e-05     \\
\textbf{measure[T.FPGD]}                   &       0.0053  &        0.000     &    10.675  &         0.000        &        0.004    &        0.006     \\
\textbf{measure[T.N400]}                   &      -0.0069  &        0.001     &   -13.720  &         0.000        &       -0.008    &       -0.006     \\
\textbf{measure[T.MAZE]}                   &      -0.0160  &        0.001     &   -28.315  &         0.000        &       -0.017    &       -0.015     \\
\textbf{method[T.tuned-lens]}              &       0.0003  &        0.000     &     1.365  &         0.172        &       -0.000    &        0.001     \\
\textbf{normalized\_layer}                 &      -0.0076  &        0.001     &   -12.601  &         0.000        &       -0.009    &       -0.006     \\
\textbf{measure[T.FPGD]:layer} &      -0.0034  &        0.001     &    -4.242  &         0.000        &       -0.005    &       -0.002     \\
\textbf{measure[T.N400]:layer} &       0.0100  &        0.001     &    14.084  &         0.000        &        0.009    &        0.011     \\
\textbf{measure[T.MAZE]:layer} &       0.0593  &        0.001     &    61.971  &         0.000        &        0.057    &        0.061     \\
\bottomrule
\end{tabular}
}
\caption{All data}
\label{tbl:coefficient_interaction_all}
\end{subtable}
\hfill

\begin{subtable}[h]{0.49\textwidth}
\centering
\scalebox{0.9}{
\begin{tabular}{lcccccc}
\toprule
                                           & \textbf{coef} & \textbf{std err} & \textbf{t} & \textbf{P$> |$t$|$} & \textbf{[0.025} & \textbf{0.975]}  \\
\midrule
\textbf{Intercept}                         &      -0.0002  &        0.000     &    -0.971  &         0.331        &       -0.001    &        0.000     \\
\textbf{stimuli[T.Federmeier+,
2007]}      &       0.0394  &        0.000     &   168.408  &         0.000        &        0.039    &        0.040     \\
\textbf{stimuli[T.Fillers]}                &       0.0011  &        0.000     &     7.379  &         0.000        &        0.001    &        0.001     \\
\textbf{stimuli[T.Hubbard+,
2019]}         &       0.0275  &        0.000     &   117.432  &         0.000        &        0.027    &        0.028     \\
\textbf{stimuli[T.Michaelov+,
2024]}       &       0.0286  &        0.000     &   122.213  &         0.000        &        0.028    &        0.029     \\
\textbf{stimuli[T.NS]}                     &      -0.0011  &        0.000     &    -6.066  &         0.000        &       -0.001    &       -0.001     \\
\textbf{stimuli[T.S\&F,2022]}              &       0.0279  &        0.000     &   119.263  &         0.000        &        0.027    &        0.028     \\
\textbf{stimuli[T.Szewczyk+,
2022]}        &       0.0321  &        0.000     &   137.295  &         0.000        &        0.032    &        0.033     \\
\textbf{stimuli[T.UCL]}                    &       0.0029  &        0.000     &    19.160  &         0.000        &        0.003    &        0.003     \\
\textbf{stimuli[T.W\&F,2012]}              &       0.0275  &        0.000     &   117.292  &         0.000        &        0.027    &        0.028     \\
\textbf{stimuli[T.ZuCO]}                   &       0.0274  &        0.000     &   165.961  &         0.000        &        0.027    &        0.028     \\
\textbf{model[T.Llama-3.1-8B]}             &    6.679e-05  &        0.000     &     0.299  &         0.765        &       -0.000    &        0.001     \\
\textbf{model[T.Qwen2.5-0.5B]}             &      -0.0002  &        0.000     &    -0.667  &         0.505        &       -0.001    &        0.000     \\
\textbf{model[T.Qwen2.5-1.5B]}             &    9.912e-05  &        0.000     &     0.424  &         0.672        &       -0.000    &        0.001     \\
\textbf{model[T.Qwen2.5-14B]}              &    3.475e-05  &        0.000     &     0.178  &         0.859        &       -0.000    &        0.000     \\
\textbf{model[T.Qwen2.5-32B]}              &   -2.221e-05  &        0.000     &    -0.123  &         0.902        &       -0.000    &        0.000     \\
\textbf{model[T.Qwen2.5-3B]}               &      -0.0003  &        0.000     &    -1.455  &         0.146        &       -0.001    &        0.000     \\
\textbf{model[T.Qwen2.5-72B]}              &      -0.0004  &        0.000     &    -2.508  &         0.012        &       -0.001    &     -9.3e-05     \\
\textbf{model[T.Qwen2.5-7B]}               &      -0.0005  &        0.000     &    -2.061  &         0.039        &       -0.001    &    -2.36e-05     \\
\textbf{model[T.gpt2]}                     &      -0.0003  &        0.000     &    -1.304  &         0.192        &       -0.001    &        0.000     \\
\textbf{model[T.gpt2-large]}               &    3.558e-05  &        0.000     &     0.199  &         0.842        &       -0.000    &        0.000     \\
\textbf{model[T.gpt2-medium]}              &      -0.0005  &        0.000     &    -2.208  &         0.027        &       -0.001    &    -6.13e-05     \\
\textbf{model[T.gpt2-xl]}                  &       0.0005  &        0.000     &     2.718  &         0.007        &        0.000    &        0.001     \\
\textbf{model[T.opt-1.3b]}                 &       0.0004  &        0.000     &     1.794  &         0.073        &     -3.3e-05    &        0.001     \\
\textbf{model[T.opt-125m]}                 &      -0.0001  &        0.000     &    -0.411  &         0.681        &       -0.001    &        0.000     \\
\textbf{model[T.opt-13b]}                  &       0.0004  &        0.000     &     1.747  &         0.081        &    -4.41e-05    &        0.001     \\
\textbf{model[T.opt-2.7b]}                 &       0.0005  &        0.000     &     2.341  &         0.019        &     8.51e-05    &        0.001     \\
\textbf{model[T.opt-30b]}                  &       0.0002  &        0.000     &     1.040  &         0.298        &       -0.000    &        0.001     \\
\textbf{model[T.opt-6.7b]}                 &       0.0001  &        0.000     &     0.669  &         0.503        &       -0.000    &        0.000     \\
\textbf{model[T.opt-66b]}                  &       0.0005  &        0.000     &     2.709  &         0.007        &        0.000    &        0.001     \\
\textbf{model[T.pythia-1.4b-deduped]}      &   -5.816e-05  &        0.000     &    -0.293  &         0.770        &       -0.000    &        0.000     \\
\textbf{model[T.pythia-12b-deduped]}       &       0.0005  &        0.000     &     2.831  &         0.005        &        0.000    &        0.001     \\
\textbf{model[T.pythia-14m]}               &      -0.0018  &        0.000     &    -4.125  &         0.000        &       -0.003    &       -0.001     \\
\textbf{model[T.pythia-160m-deduped]}      &      -0.0008  &        0.000     &    -3.296  &         0.001        &       -0.001    &       -0.000     \\
\textbf{model[T.pythia-1b-deduped]}        &      -0.0004  &        0.000     &    -1.331  &         0.183        &       -0.001    &        0.000     \\
\textbf{model[T.pythia-1b-deduped-v0]}     &       0.0002  &        0.001     &     0.294  &         0.769        &       -0.001    &        0.001     \\
\textbf{model[T.pythia-2.8b-deduped]}      &      -0.0001  &        0.000     &    -0.646  &         0.519        &       -0.000    &        0.000     \\
\textbf{model[T.pythia-31m]}               &      -0.0018  &        0.000     &    -4.200  &         0.000        &       -0.003    &       -0.001     \\
\textbf{model[T.pythia-410m-deduped]}      &      -0.0005  &        0.000     &    -2.451  &         0.014        &       -0.001    &    -9.76e-05     \\
\textbf{model[T.pythia-6.9b-deduped]}      &       0.0003  &        0.000     &     1.807  &         0.071        &    -2.81e-05    &        0.001     \\
\textbf{model[T.pythia-70m-deduped]}       &      -0.0013  &        0.000     &    -4.092  &         0.000        &       -0.002    &       -0.001     \\
\textbf{measure[T.FPGD]}                   &       0.0053  &        0.000     &    27.964  &         0.000        &        0.005    &        0.006     \\
\textbf{measure[T.N400]}                   &      -0.0301  &        0.000     &  -124.125  &         0.000        &       -0.031    &       -0.030     \\
\textbf{measure[T.MAZE]}                   &       0.0007  &        0.000     &     3.082  &         0.002        &        0.000    &        0.001     \\
\textbf{method[T.tuned-lens]}              &      -0.0003  &     8.25e-05     &    -3.448  &         0.001        &       -0.000    &       -0.000     \\
\textbf{normalized\_layer}                 &       0.0015  &        0.000     &     6.663  &         0.000        &        0.001    &        0.002     \\
\textbf{measure[T.FPGD]:layer} &      -0.0088  &        0.000     &   -29.286  &         0.000        &       -0.009    &       -0.008     \\
\textbf{measure[T.N400]:layer} &       0.0046  &        0.000     &    16.990  &         0.000        &        0.004    &        0.005     \\
\textbf{measure[T.MAZE]:layer} &       0.0030  &        0.000     &     8.430  &         0.000        &        0.002    &        0.004     \\
\bottomrule
\end{tabular}
}
\caption{Only with sentence/clause-final tokens}
\label{tbl:coefficient_interaction_last_token}
\end{subtable}
\renewcommand{\arraystretch}{1.0}
    \caption{Obtained coefficients in~\Cref{subsec:layer_and_measure}}
    \label{tbl:coefficient_interaction}
\end{table}

\begin{table}[t]
    \centering
    \tiny
    \tabcolsep=0.02cm
    \renewcommand{\arraystretch}{0.7}
    \begin{tabular}{lrrrrrr}
    \toprule
Features   & \textbf{coef} & \textbf{std err} & \textbf{t} & \textbf{P$> |$t$|$} & \textbf{[0.025} & \textbf{0.975]}  \\
\midrule
\textbf{Intercept}                     &       0.0801  &        0.430     &     0.186  &         0.852        &       -0.764    &        0.924     \\
\textbf{model[T.gpt2-large]}           &       0.1510  &        0.061     &     2.490  &         0.013        &        0.032    &        0.270     \\
\textbf{model[T.gpt2-xl]}              &       0.1647  &        0.061     &     2.716  &         0.007        &        0.046    &        0.284     \\
\textbf{model[T.opt-1.3b]}             &       0.1293  &        0.061     &     2.132  &         0.033        &        0.010    &        0.248     \\
\textbf{model[T.opt-125m]}             &       0.0182  &        0.061     &     0.301  &         0.764        &       -0.101    &        0.137     \\
\textbf{model[T.opt-6.7b]}             &       0.1898  &        0.061     &     3.131  &         0.002        &        0.071    &        0.309     \\
\textbf{model[T.pythia-1.4b-deduped]}  &       0.0754  &        0.061     &     1.243  &         0.214        &       -0.043    &        0.194     \\
\textbf{model[T.pythia-12b-deduped]}   &       0.1649  &        0.061     &     2.719  &         0.007        &        0.046    &        0.284     \\
\textbf{model[T.pythia-160m-deduped]}  &       0.1237  &        0.061     &     2.040  &         0.041        &        0.005    &        0.243     \\
\textbf{model[T.pythia-1b-deduped-v0]} &       0.0695  &        0.061     &     1.146  &         0.252        &       -0.049    &        0.188     \\
\textbf{model[T.pythia-2.8b-deduped]}  &       0.1212  &        0.061     &     1.998  &         0.046        &        0.002    &        0.240     \\
\textbf{model[T.pythia-410m-deduped]}  &       0.0607  &        0.061     &     1.001  &         0.317        &       -0.058    &        0.180     \\
\textbf{model[T.pythia-6.9b-deduped]}  &       0.1534  &        0.061     &     2.529  &         0.011        &        0.035    &        0.272     \\
\textbf{model[T.pythia-70m-deduped]}   &      -0.0478  &        0.061     &    -0.788  &         0.431        &       -0.167    &        0.071     \\
\cmidrule(lr){1-7}
\textbf{pos[T.\$]}                     &      -0.1911  &        0.776     &    -0.246  &         0.805        &       -1.711    &        1.329     \\
\textbf{pos[T.,]}                      &      -4.1266  &        0.985     &    -4.188  &         0.000        &       -6.058    &       -2.195     \\
\textbf{pos[T..]}                      &      -2.5070  &        1.327     &    -1.890  &         0.059        &       -5.107    &        0.093     \\
\textbf{pos[T.CC]}                     &      -0.7410  &        0.431     &    -1.719  &         0.086        &       -1.586    &        0.104     \\
\textbf{pos[T.CD]}                     &      -1.6381  &        0.443     &    -3.699  &         0.000        &       -2.506    &       -0.770     \\
\textbf{pos[T.DT]}                     &      -0.9570  &        0.428     &    -2.238  &         0.025        &       -1.795    &       -0.119     \\
\textbf{pos[T.EX]}                     &      -1.5530  &        0.489     &    -3.176  &         0.001        &       -2.511    &       -0.595     \\
\textbf{pos[T.FW]}                     &       4.2889  &        0.580     &     7.395  &         0.000        &        3.152    &        5.426     \\
\textbf{pos[T.IN]}                     &      -1.1206  &        0.427     &    -2.625  &         0.009        &       -1.957    &       -0.284     \\
\textbf{pos[T.JJ]}                     &      -1.4790  &        0.427     &    -3.461  &         0.001        &       -2.317    &       -0.641     \\
\textbf{pos[T.JJR]}                    &      -1.6572  &        0.463     &    -3.580  &         0.000        &       -2.565    &       -0.750     \\
\textbf{pos[T.JJS]}                    &      -2.1918  &        0.488     &    -4.489  &         0.000        &       -3.149    &       -1.235     \\
\textbf{pos[T.MD]}                     &      -1.6324  &        0.435     &    -3.749  &         0.000        &       -2.486    &       -0.779     \\
\textbf{pos[T.NN]}                     &      -2.0801  &        0.427     &    -4.876  &         0.000        &       -2.916    &       -1.244     \\
\textbf{pos[T.NNP]}                    &      -1.0809  &        0.428     &    -2.523  &         0.012        &       -1.921    &       -0.241     \\
\textbf{pos[T.NNPS]}                   &      -5.2269  &        0.484     &   -10.804  &         0.000        &       -6.175    &       -4.279     \\
\textbf{pos[T.NNS]}                    &      -2.1715  &        0.428     &    -5.071  &         0.000        &       -3.011    &       -1.332     \\
\textbf{pos[T.PDT]}                    &      -2.2381  &        0.535     &    -4.183  &         0.000        &       -3.287    &       -1.189     \\
\textbf{pos[T.POS]}                    &      -2.0896  &        0.449     &    -4.651  &         0.000        &       -2.970    &       -1.209     \\
\textbf{pos[T.PRP]}                    &      -1.0883  &        0.430     &    -2.529  &         0.011        &       -1.932    &       -0.245     \\
\textbf{pos[T.PRP\$]}                  &      -1.4344  &        0.435     &    -3.296  &         0.001        &       -2.287    &       -0.581     \\
\textbf{pos[T.RB]}                     &      -1.7242  &        0.428     &    -4.026  &         0.000        &       -2.564    &       -0.885     \\
\textbf{pos[T.RBR]}                    &      -1.8143  &        0.482     &    -3.761  &         0.000        &       -2.760    &       -0.869     \\
\textbf{pos[T.RBS]}                    &      -3.0363  &        0.529     &    -5.743  &         0.000        &       -4.073    &       -2.000     \\
\textbf{pos[T.RP]}                     &      -1.4753  &        0.456     &    -3.235  &         0.001        &       -2.369    &       -0.582     \\
\textbf{pos[T.SYM]}                    &       2.6044  &        0.901     &     2.889  &         0.004        &        0.838    &        4.371     \\
\textbf{pos[T.TO]}                     &      -0.6908  &        0.431     &    -1.602  &         0.109        &       -1.536    &        0.155     \\
\textbf{pos[T.UH]}                     &      -7.8480  &        0.759     &   -10.338  &         0.000        &       -9.336    &       -6.360     \\
\textbf{pos[T.VB]}                     &      -1.2679  &        0.429     &    -2.955  &         0.003        &       -2.109    &       -0.427     \\
\textbf{pos[T.VBD]}                    &      -1.6735  &        0.431     &    -3.885  &         0.000        &       -2.518    &       -0.829     \\
\textbf{pos[T.VBG]}                    &      -2.0877  &        0.433     &    -4.818  &         0.000        &       -2.937    &       -1.238     \\
\textbf{pos[T.VBN]}                    &      -1.7750  &        0.431     &    -4.118  &         0.000        &       -2.620    &       -0.930     \\
\textbf{pos[T.VBP]}                    &      -1.1485  &        0.432     &    -2.659  &         0.008        &       -1.995    &       -0.302     \\
\textbf{pos[T.VBZ]}                    &      -1.2311  &        0.431     &    -2.860  &         0.004        &       -2.075    &       -0.387     \\
\textbf{pos[T.WDT]}                    &      -1.2174  &        0.449     &    -2.714  &         0.007        &       -2.097    &       -0.338     \\
\textbf{pos[T.WP]}                     &      -1.0289  &        0.455     &    -2.264  &         0.024        &       -1.920    &       -0.138     \\
\textbf{pos[T.WP\$]}                   &       0.1774  &        0.869     &     0.204  &         0.838        &       -1.525    &        1.880     \\
\textbf{pos[T.WRB]}                    &      -0.8454  &        0.456     &    -1.853  &         0.064        &       -1.740    &        0.049     \\
\textbf{has\_punct[T.True]}            &      -1.6396  &        0.038     &   -43.221  &         0.000        &       -1.714    &       -1.565     \\
\textbf{has\_num[T.True]}              &       2.3217  &        0.171     &    13.602  &         0.000        &        1.987    &        2.656     \\
\textbf{freq}              &       0.0142  &        0.007     &     1.930  &         0.054        &       -0.000    &        0.029     \\
\textbf{length}                        &       0.3629  &        0.007     &    51.621  &         0.000        &        0.349    &        0.377     \\
\textbf{token\_position\_in\_sentence}              &      -0.0042  &        0.001     &    -3.958  &         0.000        &       -0.006    &       -0.002     \\
\bottomrule
    \end{tabular}
    \renewcommand{\arraystretch}{1.0}
    \caption{Obtained coefficients in~\Cref{subsec:context}. Note that word frequency and length are confounded (Pearson's $r=-0.7$), and once the length factor is excluded, the coefficient for the word frequency feature becomes significantly negative and has a high t-score. That is, a large decrease in regression error is generally associated with infrequent, long words.}
    \label{tab:error_regression}
\end{table}

\begin{table}[ht]
    \centering
    \tiny
    \tabcolsep=0.01cm
    \renewcommand{\arraystretch}{0.5}
    \begin{tabular}{lrrrrrp{0.1cm}rrrrr}
        \toprule
 & \multicolumn{5}{c}{Logit-lens ($\Delta$LL)} && \multicolumn{5}{c}{Tuned-lens ($\Delta$LL)} \\
 \cmidrule(lr){2-6}  \cmidrule(lr){8-12}
Measure & 0-0.2 & 0.2-0.4 & 0.4-0.6 & 0.6-0.8 & 0.8-1 & & 0-0.2 & 0.2-0.4 & 0.4-0.6 & 0.6-0.8 & 0.8-1 \\
    \cmidrule(r){1-1} \cmidrule(lr){2-2} \cmidrule(lr){3-3} \cmidrule(lr){4-4} \cmidrule(lr){5-5} \cmidrule(lr){6-6} \cmidrule(lr){8-8}   \cmidrule(lr){9-9}  \cmidrule(lr){10-10}  \cmidrule(lr){11-11} \cmidrule(lr){12-12} 
SPR & \textbf{24.51} & 23.35 & 18.84 & 7.80 & 1.81 && \textbf{15.78} & 8.92 & 4.87 & 2.53 & 1.27 \\
FPGD  & 22.11 & \textbf{23.39} & 22.83 & 15.77 & 6.53 & & \textbf{16.28} & 14.48 & 11.87 & 9.47 & 5.57 \\
SPGD & \textbf{228.29} & 189.49 & 144.70 & 72.48 & 21.47 & & \textbf{68.43} & 20.05 & 5.28 & 1.71 & 3.87 \\
TOTAL & \textbf{201.36} & 171.44 & 134.24 & 68.64 & 19.82 & & \textbf{67.39} & 23.51 & 7.82 & 1.61 & 1.56\\
ELAN (125--175ms) & \textbf{0.75} & 0.48 & 0.31 & 0.36 & 0.34 & & 0.22 & 0.38 & \textbf{0.77} & 0.67 & 0.50 \\
LAN (300--400ms) & \textbf{67.55} & 51.86 & 35.96 & 17.45 & 8.17 & & \textbf{19.94} & 2.74 & 1.72 & 5.02 & 8.76  \\
N400 (300--500ms) & \textbf{56.86} & 38.04 & 22.77 & 16.07 & 22.58 & & 11.31 & 6.12 & 16.19 & 29.49 & \textbf{37.11} \\
EPNP (400--600ms) & \textbf{75.94} & 61.74 & 46.02 & 22.85 & 7.84 & & \textbf{29.31} & 6.67 & 1.29 & 1.56 & 4.45 \\
P600 (500--700ms) & \textbf{69.02} & 55.91 & 41.88 & 22.60 & 10.37 & & \textbf{17.70} & 3.30 & 1.80 & 5.83 & 11.36  \\
PNP (600--700ms) & \textbf{59.75} & 50.83 & 39.67 & 21.40 & 7.26 & & \textbf{25.38} & 8.09 & 2.30 & 0.73 & 1.61  \\
        \bottomrule
        \end{tabular}
        \renewcommand{\arraystretch}{1.0}
        \caption{Results for other human measures recorded on the UCL corpus. $\Delta$LLs are multiplied by 1000.} 
        \label{tbl:results_other_variables}
\end{table}

\begin{figure*}[ht]
    \centering
    \begin{subfigure}[t]{0.49\textwidth}
        \includegraphics[width=1.0\linewidth]{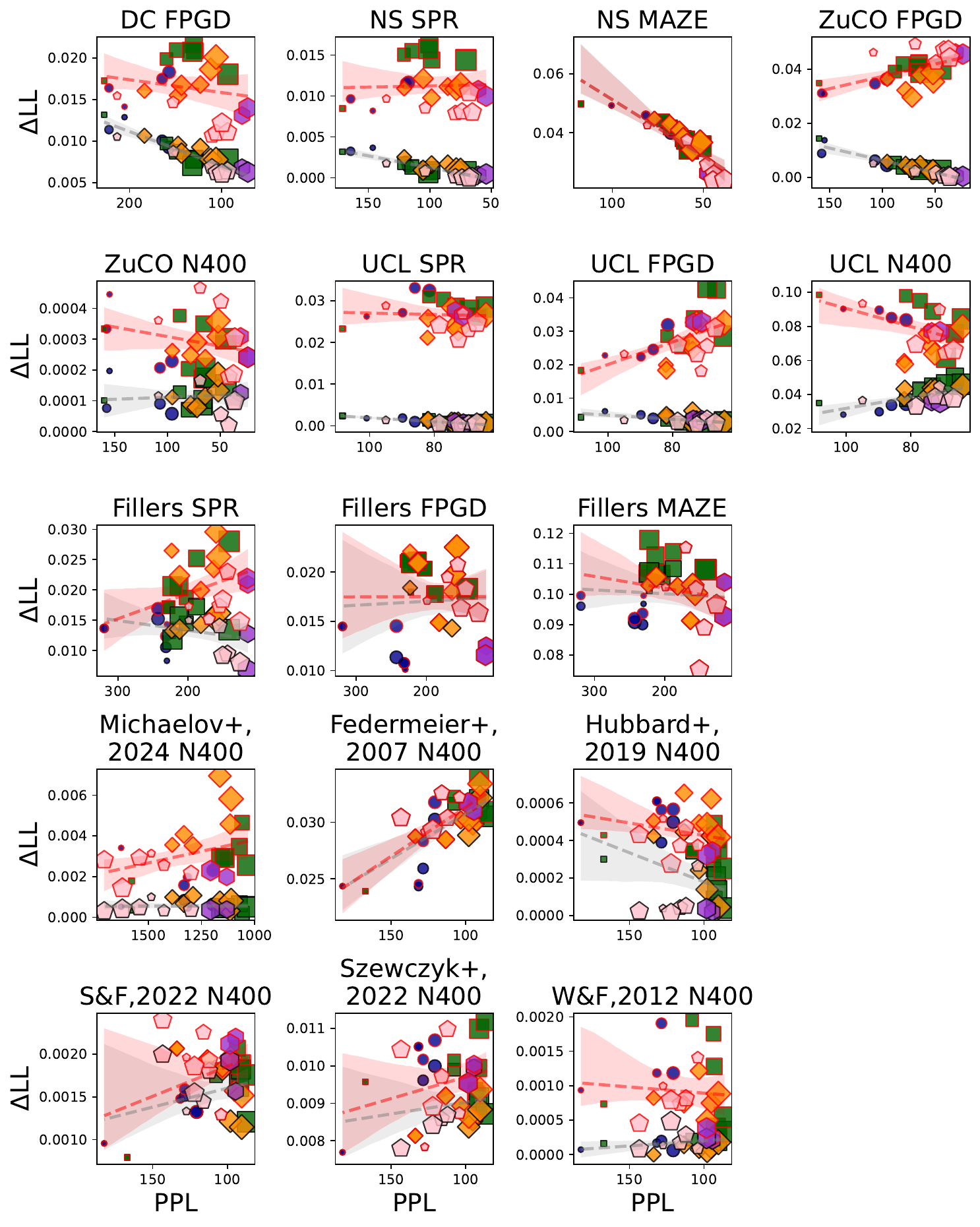}
        \caption{Logit-lens}
    \end{subfigure}
    \hfill
        \begin{subfigure}[t]{0.49\textwidth}
        \includegraphics[width=1.0\linewidth]{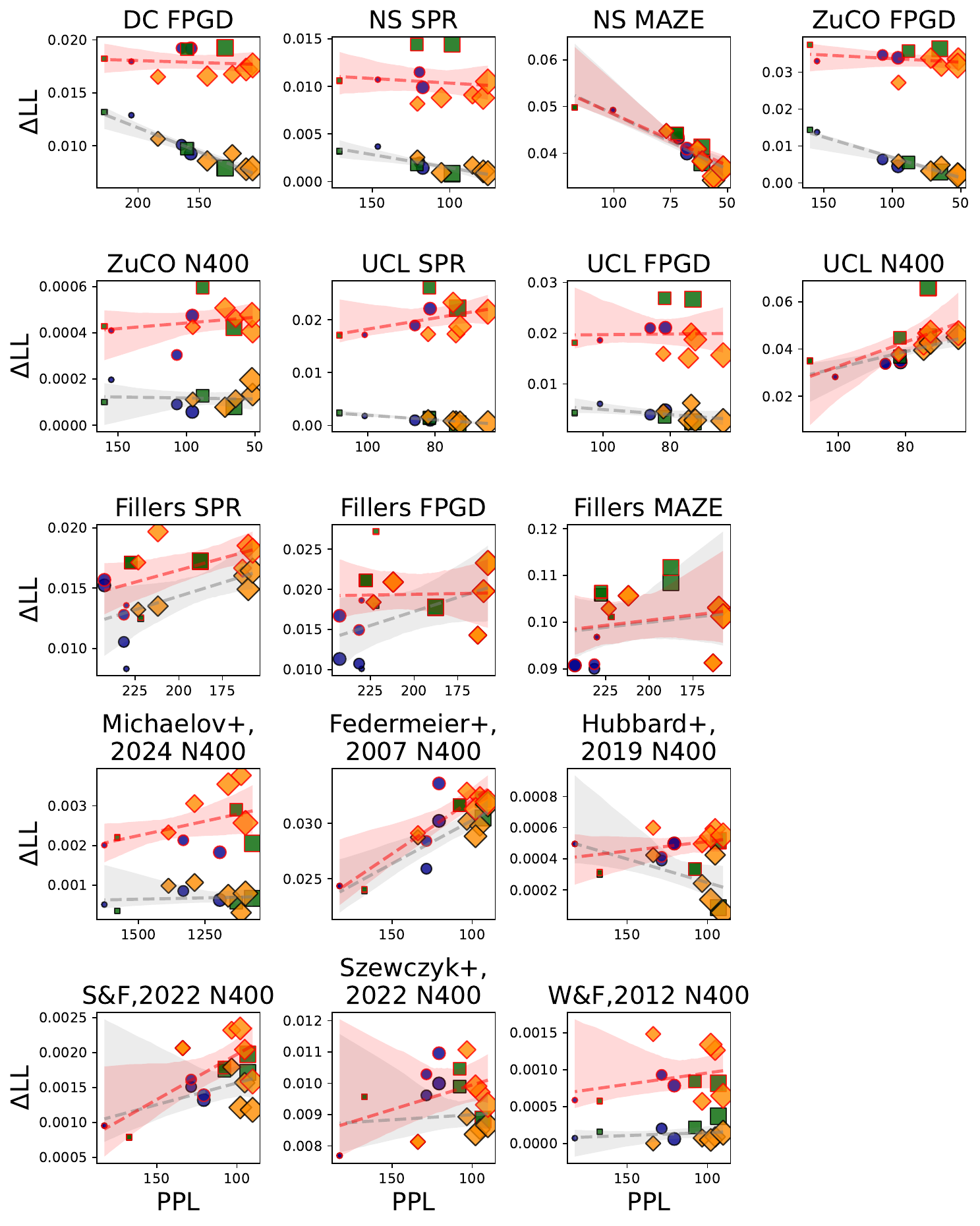}
        \caption{Tuned-lens}
    \end{subfigure}
    \caption{
Scaling effect between $\Delta$LL and PPL (measured on respective datasets with final layer), instead of model parameter counts, as adopted in Figure~\ref{fig:params_ppp}.
    }
    \label{fig:ppl_ppp}
\end{figure*}

\clearpage

\begin{figure*}
    \centering
    \includegraphics[width=\linewidth]{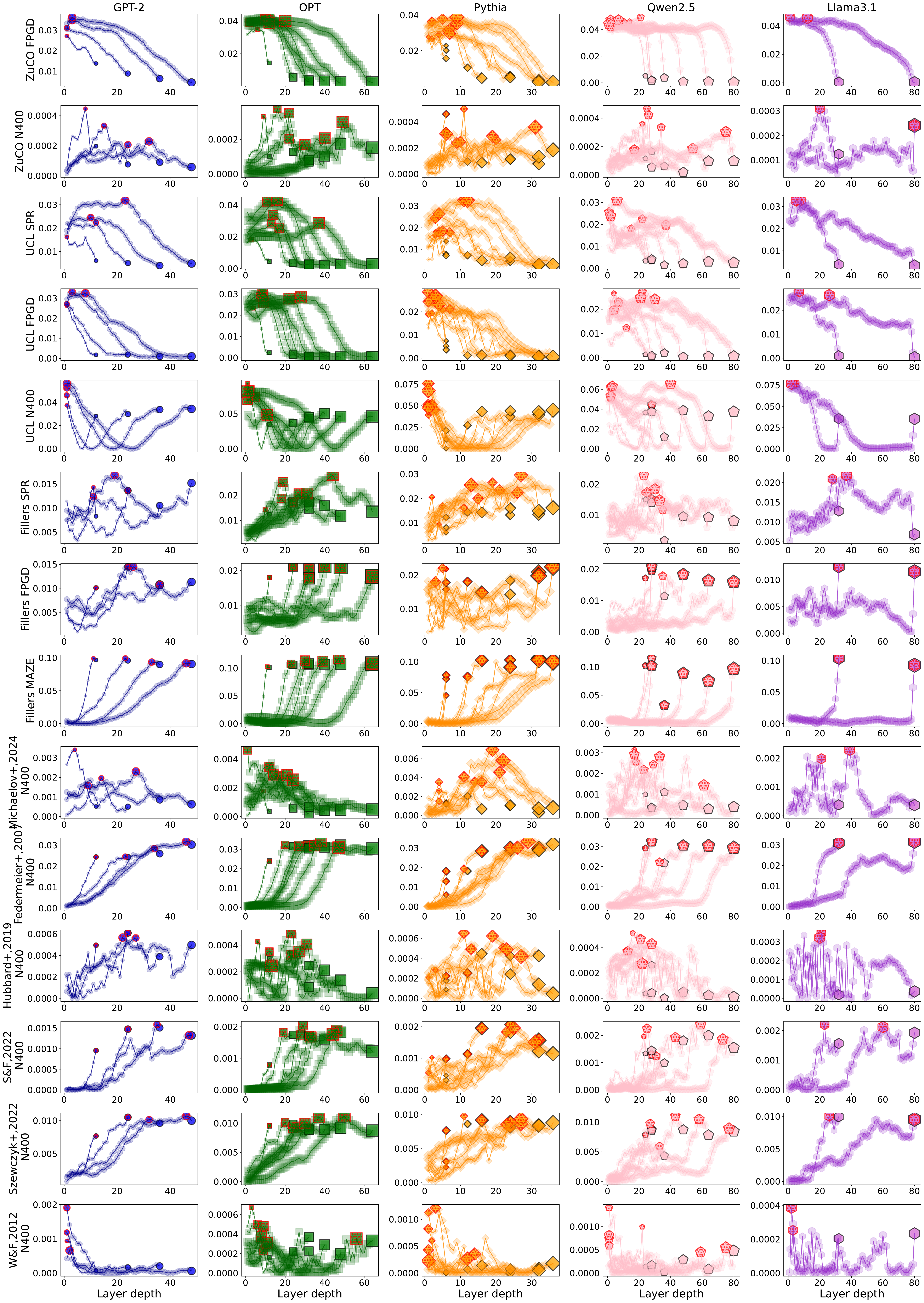}
    \caption{Visulaition of layer--$\Delta$LL relationships (in addition to Figure~\ref{fig:overview})}
    \label{fig:enter-label}
\end{figure*}

\end{document}